\documentclass{article} 
\usepackage{iclr2027_conference,times}

\usepackage{amsmath,amsfonts,bm}

\def\eqref#1{equation~\ref{#1}}

\def\1{\bm{1}}

\DeclareMathAlphabet{\mathsfit}{\encodingdefault}{\sfdefault}{m}{sl}
\SetMathAlphabet{\mathsfit}{bold}{\encodingdefault}{\sfdefault}{bx}{n}

\usepackage{hyperref}
\usepackage{url}

\usepackage{graphicx}
\usepackage[flushleft]{threeparttable}
\usepackage{multirow}
\usepackage{subcaption}
\usepackage{wrapfig}

\usepackage{float}
\usepackage{algorithm}
\usepackage{algorithmic}  
\usepackage{enumitem}

\usepackage{amsmath}
\usepackage{amssymb}
\usepackage{mathtools}
\usepackage{amsthm}
\usepackage{bm}
\usepackage{amsfonts}       
\usepackage{svg}
\usepackage[capitalize,noabbrev]{cleveref}
\usepackage{rotating}

\usepackage{microtype}      
\usepackage{titletoc}
\usepackage{longtable}
\usepackage{booktabs}       
\usepackage{tabularx}
\usepackage{xcolor}         
\usepackage[table]{xcolor} 
\usepackage{CJKutf8} 
\newcommand{\greybg}[1]{{\cellcolor[HTML]{d6e8d4}\textbf{#1}}} 

\newsavebox{\heatmapgrid}

\title{MiLoop: Selective Memory Propagation for \\ Neural Combinatorial Optimization}

\author{
Changliang Zhou$^{1,2*}$,
Yuanyao Chen$^{1,2}$\thanks{Equal contribution},
Rongsheng Chen$^{1,2,3}$,
Zhiyun Lin$^{1,2}$,
Zhenkun Wang$^{1,2}$\thanks{Corresponding author}\\
$^1$School of Automation and Intelligent Manufacturing, \\
\ \ Southern University of Science and Technology, Shenzhen, China \\
$^2$Guangdong Provincial Key Laboratory of Fully Actuated System Control Theory and Technology,\\
\ \ Southern University of Science and Technology, Shenzhen, China \\
$^3$ Pengcheng Laboratory, Shenzhen, China\\
  \texttt{\{zhoucl2022,12433020,chenrs2025\}@mail.sustech.edu.cn}  \\
\texttt{\{linzy, wangzk3\}@sustech.edu.cn} \\
} 

\iclrfinalcopy 
\begin{document}
\begin{CJK*}{UTF8}{gbsn} 

\maketitle

\begin{abstract}

Constructive neural combinatorial optimization (NCO) has emerged as a promising paradigm that learns to construct solutions to combinatorial optimization problems (COPs) step by step, which reduces reliance on handcrafted rules and enables fast inference. While many methods with dynamic embeddings generalize well, they typically rebuild subproblem representations from scratch at each step using deep attention stacks. Many high-performing methods in this category rely on solution labels or pseudo-labels for efficient training, or on aggressive search space pruning during reinforcement learning (RL). To address these limitations, we propose \textbf{Memory-in-the-Loop (MiLoop)}, a purely RL-based constructive framework that leverages the multi-step computation already required by a rollout for selective memory propagation. Each rollout provides solution-quality feedback for learning while propagating historical representations, thereby enabling a shallow policy to learn effective dynamic embeddings without external solution labels or training-time search-space pruning. Specifically, MiLoop fuses current embeddings with historical memory before the attention layers and applies adaptive gated updates afterward. The updated representations support both current decisions and stepwise reuse. Extensive experiments across four COPs demonstrate that MiLoop consistently produces high-quality solutions on instances \textbf{ranging from 100 to 10 million nodes}, highlighting its strong generalization ability.
\end{abstract}

\section{Introduction}
\label{sec:Introduction}
Combinatorial optimization problems (COPs) have extensive applications across various domains, including supply chain management~\citep{guerriero2025supplychain}, chip design~\citep{gao2026eda}, and drug discovery~\citep{liu2017combinatorial_drug}. Many practically important COPs are NP-hard, making exact optimization computationally demanding as the instance size increases. Over the past decades, classical heuristics have achieved competitive solution quality within acceptable time budgets~\citep{HGS,LKH3}. However, they rely on handcrafted rules that require substantial domain expertise and problem-specific adaptation. This limitation increases the effort required to develop solvers for complex COPs, especially when expert knowledge is unavailable or expensive to acquire.

In recent years, neural combinatorial optimization (NCO) has emerged as an alternative that reduces reliance on handcrafted rules by learning heuristic rules from data~\citep{bengio2021machine,ba2026survey}. Among existing NCO methods, constructive neural models directly generate complete feasible solutions for a given instance in a single rollout, without requiring time-consuming iterative search. Their efficient inference makes them particularly attractive when facing real-time scenarios. Existing constructive methods generally follow either an autoregressive (AR)~\citep{vinyals2015pointer,kool2019attention,zhou2025L2R} or a non-autoregressive (NAR)~\citep{joshi2019efficient,sun2023difusco} paradigm. Compared with the NAR construction, AR construction~\citep{kwon2020pomo,gao2023elg,zhou2026urs} enables the model to adjust its decisions as the generated partial solution state evolves and is readily applicable to a wide range of COPs. These properties have made AR construction one of the predominant paradigms in NCO.

\begin{figure}[t]
    \centering
    \includegraphics[width=\linewidth]{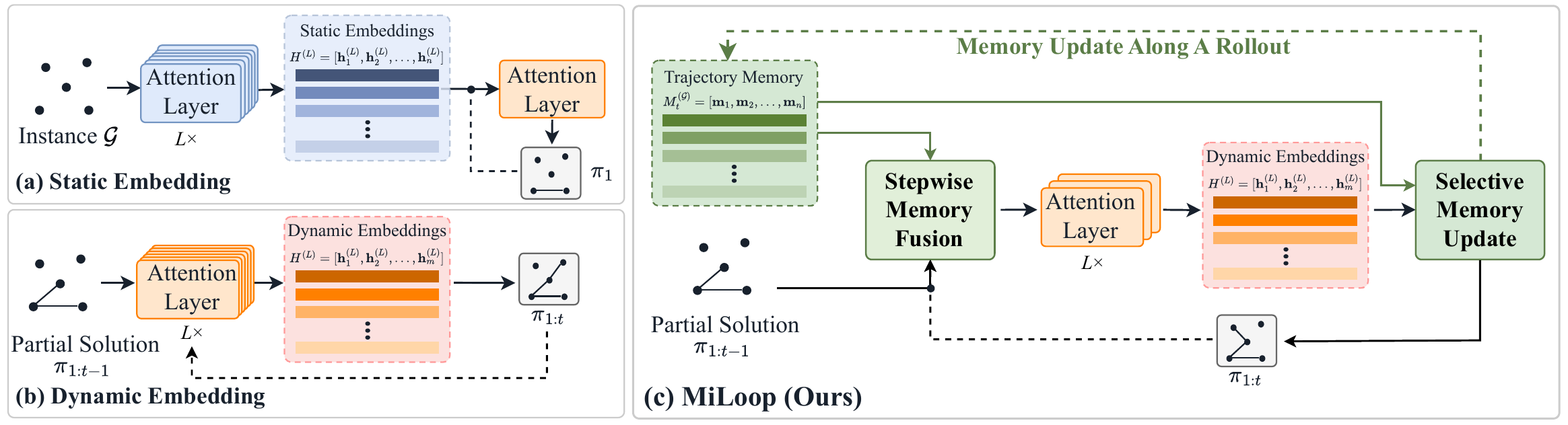}
    \caption{Comparison of different embeddings. Unlike traditional dynamic embeddings, MiLoop maintains a trajectory memory throughout each rollout, with selective updates for subsequent reuse.}
    \label{fig:embedding_comparison}
\vspace{-15pt}
\end{figure}

As shown in \cref{fig:embedding_comparison}, AR constructive neural solvers broadly adopt two embedding strategies: \emph{static} and \emph{dynamic}. Methods with static embeddings~\citep{kwon2020pomo,zhou2026icam} compute node or edge embeddings once and reuse them across subsequent decoding steps. Their computational efficiency and compatibility with reinforcement learning (RL) have contributed to their widespread adoption. In contrast, methods with dynamic embeddings~\citep{luo2023lehd,drakulic2023bq} re-embed the current subproblem at each construction step, producing representations that reflect the current construction state. These methods have shown promising generalization to larger instances.

Despite these advances, each embedding strategy faces its own challenges. Static-embedding methods still exhibit substantial degradation in solution quality on instances much larger than those seen during training (e.g., with more than 1,000 nodes)~\citep{kwon2020pomo,kim2022symnco}. Many dynamic-embedding methods reconstruct contextual relationships within the current subproblem from scratch at each step, without reusing representations computed in previous steps. They typically use relatively deep Transformer~\citep{vaswani2017attention} stacks (e.g., 6 layers) to obtain effective embeddings~\citep{luo2023lehd,drakulic2023bq}. Many high-performing methods in this category are trained using supervised learning~\citep{luo2023lehd,drakulic2023bq} or self-improvement with high-quality pseudo-labels generated by advanced search techniques~\citep{luo2024Boosting,pirnay2024self}. More recently, search space reduction has been used to ease solution construction in RL-trained solvers with dynamic embeddings~\citep{fang2024invit,zhou2025L2R}. However, aggressive pruning during training may exclude actions required for high-quality solutions. A detailed review of constructive methods with static and dynamic embeddings is provided in Appendix \ref{append:related_work}.

Recent advances in large language models have demonstrated the potential of \emph{looped Transformers} for reasoning~\citep{saunshi2025reasoning,jeddi2026loopformer}. These models repeatedly apply shared Transformer blocks to refine latent representations, increasing effective depth without increasing the parameter count. In RL-based constructive NCO methods, each rollout already involves repeatedly applying the same policy to construct a complete solution. This sequential structure provides a natural basis for a memory loop across construction steps. However, as the construction state changes, the relevance of historical information may vary across nodes, feature channels, and construction steps. Effective memory reuse requires incorporating current information while retaining and revising useful historical context. This raises the following question: \textit{Can we turn the multi-step computation already required by RL rollouts into an advantage through selective memory propagation, enabling a shallow policy to learn effective dynamic embeddings?}

In this paper, we propose \textbf{\underline{M}}emory-\textbf{\underline{i}}n-the-\textbf{\underline{Loop}} (MiLoop), an RL-based framework that integrates selective memory propagation into solution construction rollouts. Each rollout provides solution-quality feedback for learning while carrying structural representations forward to subsequent decisions. By reusing memories across decoding steps, MiLoop uses substantially fewer attention layers than the existing dynamic-embedding baselines. Our main contributions are as follows:
\begin{itemize}
    \item We introduce an RL-based framework that leverages the multi-step computation already required by construction rollouts to enable stepwise representation reuse, allowing a shallow policy to learn dynamic embeddings without any external solution labels.

    \item We develop a selective memory propagation architecture that fuses current embeddings with memory states before the attention stacks and applies adaptive gated updates afterward at each construction step, allowing the updated representations to support both current decisions and reuse in subsequent steps.

    \item Our evaluation covers four COPs and includes instances ranging from 100 to 10 million nodes. Extensive experiments demonstrate that MiLoop consistently produces high-quality solutions across this size range and exhibits strong generalization ability.
\end{itemize}

\section{Preliminaries}
\label{sec:preliminaries}

\subsection{Constructive Neural Solver with Dynamic Embeddings}
\label{sec:decoder_only_construction}
Given a COP instance $\mathcal G$ of size $N$, let $\Omega(\mathcal G)$ denote its discrete feasible solution set and $f(\bm y\mid\mathcal G)$ denote the objective function of solution $\bm y$. For consistency, we formulate all problems as minimization problems, converting maximization objectives by negating them, i.e., $\bm y^\star
\in
\arg\min_{\bm{y} \in \Omega(\mathcal{G})}
f(\bm{y} \mid \mathcal{G}).$

The constructive methods with dynamic embeddings have attracted interest for their promising cross-scale generalization, which recomputes node representations from the current subproblem $\mathcal{G}_{t}$ at each construction step~\citep{drakulic2023bq,luo2023lehd}. Without loss of generality, we use Traveling Salesman Problem (TSP) to illustrate the general pipeline and denote the learnable parameters by $\bm\theta=\{\bm\theta_{\mathrm{init}},\bm\theta_{\mathrm{attn}},\bm\theta_{\mathrm{logit}}\}$, corresponding to initial embedding layer, attention layers, and output scoring layer, respectively. 

For TSP, the starting node $\pi_1$ is randomly selected at the first construction step. At step $t\geq 2$, given the partial solution $\pi_{1:t-1}$, the model input consists of the starting node $\pi_1$, the current node $\pi_{t-1}$, and a node set $\mathcal A_t$ containing the unvisited nodes in $\mathcal G$. For each node $i\in\{\pi_1,\pi_{t-1}\}\cup\mathcal A_t$, its feature $\bm v_{t,i}\in\mathbb R^{1\times2}$ is mapped by $\bm\theta_{\mathrm{init}}$ to a $d$-dimensional embedding. This yields the initial representation of the current subproblem, $H_t^{(0)}=\{\bar{\mathbf{h}}_{\pi_1}^{(0)}, \bar{\mathbf{h}}_{\pi_{t-1}}^{(0)}\} \cup H_{\mathcal{A}_{t}}^{(0)}\in \mathbb{R}^{(2+|\mathcal{A}_{t}|) \times d}$, where $\bar{\mathbf{h}}_{\pi_{1}}^{(0)}$ and $\bar{\mathbf{h}}_{\pi_{t-1}}^{(0)}$ denote the role-specific embeddings of $\pi_1$ and $\pi_{t-1}$, respectively. The $H_t^{(0)}$ is then fed into $\bm\theta_{\mathrm{attn}}$ with $L$ attention layers~\citep{vaswani2017attention}. For layer $\ell$, the output for node $i$ is computed as
\begin{equation}
\begin{aligned}
\hat{\mathbf{h}}_{t,i}^{(\ell)} &= \operatorname{Norm}_{\ell,1}\left(\mathbf{h}_{t,i}^{(\ell-1)} + \operatorname{Attention}_{\ell}\left(\mathbf{h}_{t,i}^{(\ell-1)}, H_{t}^{(\ell-1)}\right)\right), \\
\mathbf{h}_{t,i}^{(\ell)} &= \operatorname{Norm}_{\ell,2}\left(\mathbf{\hat{h}}_{t,i}^{(\ell)} + \operatorname{FFN}_{\ell}\left(\mathbf{\hat{h}}_{t,i}^{(\ell)}\right)\right), \qquad i = 1,2,\ldots,2+|\mathcal A_t|
\end{aligned}
\label{eq:attention_layer}
\end{equation}
where $\operatorname{Attention}(\cdot)$ means the adopted attention mechanism and $\operatorname{FFN}(\cdot)$ indicates a fully connected neural network, both equipped with residual connections~\citep{he2016skip}. Note that $\operatorname{Norm}(\cdot)$ is optional~\citep{luo2023lehd}. We denote the complete stack by $H_t^{(L)}
=
\operatorname{AttnLayers}_{\bm\theta_{\mathrm{attn}}}^{(L)}
\left(H_t^{(0)}\right),$
where $H_t^{(L)}$ contains the advanced embeddings of each node. Finally, an output module maps $H_t^{(L)}$ to candidate logits, using either a linear projection~\citep{drakulic2023bq} or a compatibility calculation~\citep{zhou2025L2R}. Denoting this module by $\operatorname{ActionSelection}_{\bm\theta_{\mathrm{logit}}}$, we obtain $\bm u_t
=
\operatorname{ActionSelection}_{\bm\theta_{\mathrm{logit}}}
\left(H_t^{(L)}\right)
\in\mathbb R^{|\mathcal A_t|},$
where $u_{t,i}$ is the logit for each feasible node $i$ at step $t$.

\subsection{Autoregressive Construction}
\label{sec:autoregressive_construction}
An autoregressive constructive solver builds a solution through a finite action sequence $\bm\pi=(\pi_1,\ldots,\pi_T)$, conditioning each decision on the current partial solution $\pi_{1:t-1}$ and constraint state $s_t$. At step $t$, let $\mathcal H_t=(\mathcal G,\pi_{1:t-1},s_t)$ denote the available decision history, and let $\mathcal N_t\subseteq\mathcal A_t$ denote the set of feasible actions. For any $j \in\mathcal A_t\setminus\mathcal N_t$, its logit $u_{\bm\theta}(j,\mathcal H_t)$ is masked to $-\infty$. Given the logit $u_{\bm\theta}(a,\mathcal H_t)$ for action $i$, the masked action distribution is
\begin{equation}
\label{eq:masked_policy}
p_{\bm\theta}(\pi_t=i\mid\mathcal H_t)
=
\frac{\exp u_{\bm\theta}(i,\mathcal H_t)}
{\sum_{j\in\mathcal A_t}\exp u_{\bm\theta}(j,\mathcal H_t)},
\quad i\in\mathcal A_t,
\end{equation}
Selecting $\pi_t$ extends the partial solution to $\pi_{1:t}$ and updates the constraint state. This process continues until a problem-specific termination condition is met. The probability of the complete solution $\bm \pi$ factorizes as
\begin{equation}
\label{eq:ar_factorization}
p_{\bm\theta}(\bm\pi\mid\mathcal G)
=
\prod_{t=1}^{T}
p_{\bm\theta}
\left(
\pi_t\mid\mathcal G,\pi_{1:t-1},s_t
\right).
\end{equation}
Most AR solvers with dynamic embeddings recompute contextualized representations for the nodes in $\mathcal A_t$ at each step. While this allows them to adapt to the current problem state, it does not retain previously computed representations for reuse in subsequent steps. Under an AR formulation, solving a COP can be viewed as a sequential decision process, which provides a natural basis for memory propagation, allowing neural solvers to retain and reuse representations across construction steps.

\section{Methodology}
\label{sec:methodology}

\begin{figure}[t]
    \centering
    \includegraphics[width=\linewidth]{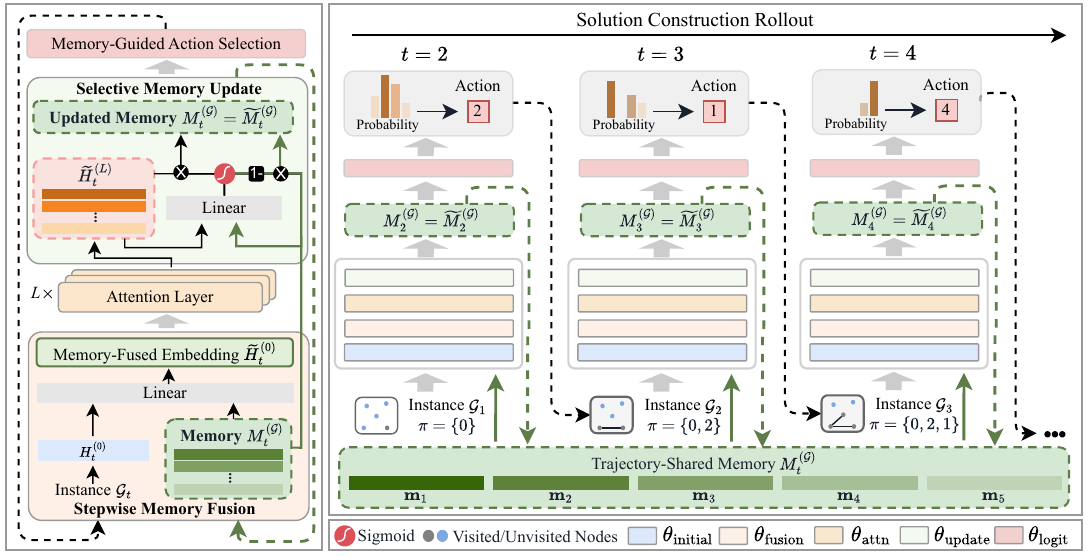}
    \caption{Overview of MiLoop, illustrated on the TSP instances, the starting node $\pi_1$ is randomly selected at the first step. During RL rollouts, MiLoop fuses node-wise memory with current embeddings before attention and selectively updates it afterward for action scoring and cross-step reuse.}
    \label{fig:methodology}
    \vspace{-10pt}
\end{figure}
Unlike existing neural solvers, MiLoop integrates selective memory propagation into rollout-based RL training, allowing the policy to reuse previously computed representations across its decision sequence. The full parameter set is $\bm\theta=\{\bm\theta_{\mathrm{init}},\bm\theta_{\mathrm{fusion}},\bm\theta_{\mathrm{attn}},\bm\theta_{\mathrm{update}},\bm\theta_{\mathrm{logit}}\}$, corresponding to input embedding, memory fusion, attention layers, memory updating, and action scoring, respectively. 

As illustrated in Figure~\ref{fig:methodology}, MiLoop maintains a global node-wise memory matrix $M_t^{(\mathcal{G})}=[\mathbf{m}_{1},\mathbf{m}_{2},\ldots,\mathbf{m}_{n}]\in\mathbb R^{N\times d}$, whose $i$-th row $\mathbf m_{t,i}\in\mathbb R^{1\times d}$ represents the memory state of node $i$ at the start of step $t$. Specifically, MiLoop fuses current inputs with historical memory via $\bm\theta_{\mathrm{fusion}}$ prior to feed into $\bm\theta_{\mathrm{attn}}$ and applies adaptive gates $\bm\theta_{\mathrm{update}}$ to the refined current embeddings to obtain a new memory matrix $M_{t+1}^{(\mathcal G)}$, which is further used for current action scoring and is carried forward to subsequent steps, allowing the same shallow decoder to reuse and revise memory throughout the rollout. We describe the architecture using TSP as an example and provide task-specific implementations for other evaluated COPs in Appendix~\ref{append:implementation_details}.

\subsection{Stepwise Memory Fusion}
\label{sec:stepwise_memory_fusion}
Although the memory matrix $M_t^{(\mathcal G)}$ available at the start of step $t$ retains context from earlier construction states, this context alone may not fully reflect the current input. MiLoop therefore fuses current input embeddings with historical memory before feeding into $\bm\theta_{\mathrm{attn}}$ at each step, allowing its shallow decoder to process the current subproblem while reusing previously computed embeddings.

\paragraph{Initial Embedding} 
At the construction step $t\geq 2$, we introduce a coordinate normalization operation~\citep{fang2024invit,zhou2025L2R} to ensure that each subproblem $\mathcal G_{t}$ adheres to a similar distribution, which simplifies the input feature space but also significantly enhances the stability of model inputs across different sub-graphs. Then we map the features $\bm v_i\in\mathbb R^{1\times 2}$ of each node $i\in\{\pi_1,\pi_{t-1}\}\cup\mathcal A_t$ to a $d$-dimensional embedding through a shared linear projection, $\mathbf h_i^{(0)}=\bm v_iW^{(0)}+\bm b^{(0)}$, where $W^{(0)}\in\mathbb R^{2\times d}$ and $\bm b^{(0)}\in\mathbb R^{1\times d}$ are learnable parameters. To distinguish the roles of the starting node and current node, we further apply separate linear projections to their embeddings, yielding $\bar{\mathbf h}_{\pi_1}^{(0)}=\mathbf h_{\pi_1}^{(0)}W_{\mathrm{first}}^{(0)}$ and $\bar{\mathbf h}_{\pi_{t-1}}^{(0)}=\mathbf h_{\pi_{t-1}}^{(0)}W_{\mathrm{last}}^{(0)}$, with $W_{\mathrm{first}}^{(0)},W_{\mathrm{last}}^{(0)}\in\mathbb R^{d\times d}$. The initial embeddings are expressed as $H_t^{(0)}=\{\bar{\mathbf{h}}_{\pi_1}^{(0)}, \bar{\mathbf{h}}_{\pi_{t-1}}^{(0)}\} \cup H_{\mathcal{A}_{t}}^{(0)}\in \mathbb{R}^{(2+|\mathcal{A}_{t}|) \times d}$.

\paragraph{Memory-Fused Embedding} 
Let $\mathbf{m}_{\pi_1},\mathbf{m}_{\pi_{t-1}}\in\mathbb R^{1\times d}$ denote the historical representations retrieved from $M_t^{(\mathcal G)}$ for the starting node and current node, respectively. The matrix $M_{\mathcal A_t}^{(\mathcal G)}\in\mathbb R^{|\mathcal A_t|\times d}$ contains the historical representations of unvisited nodes. At the first call, we directly use $\widetilde H_t^{(0)}=H_t^{(0)}$ and use the corresponding current embeddings as the initial memories. At subsequent construction steps ($t\geq 2$), we fuse historical with current embeddings through three separate linear projections:
\begin{equation}
\label{eq:miloop_input_fusion}
\widetilde{\mathbf{h}}_{\pi_1}^{(0)}
=
\left[\mathbf{m}_{\pi_1}\Vert\bar{\mathbf{h}}_{\pi_1}^{(0)}\right]W_{\mathrm{fusion}}^{\mathrm{1}}, \quad
\widetilde{\mathbf{h}}_{\pi_{t-1}}^{(0)}
=
\left[\mathbf{m}_{\pi_{t-1}}\Vert\bar{\mathbf{h}}_{\pi_{t-1}}^{(0)}\right]W_{\mathrm{fusion}}^{\mathrm{2}},\quad
\widetilde H_{\mathcal A_t}^{(0)}
=
\left[M_{\mathcal A_t}^{(\mathcal{G})}\Vert H_{\mathcal A_t}^{(0)}\right]W_{\mathrm{fusion}}^{\mathrm{3}}
\end{equation}
where $\Vert$ denotes concatenation along the feature dimension, and $W_{\mathrm{fusion}}^{1},W_{\mathrm{fusion}}^{2},W_{\mathrm{fusion}}^{3}\in\mathbb R^{2d\times d}$ are learnable fusion matrices for the starting, current, and unvisited nodes, respectively. This fusion provides a direct path for current features into attention while allowing the decoder to reuse structural context from previous steps. The fused embeddings $\widetilde H_t^{(0)}\{\widetilde{\mathbf{h}}_{\pi_1}^{(0)}, \widetilde{\mathbf{h}}_{\pi_{t-1}}^{(0)}\} \cup \widetilde H_{\mathcal A_t}^{(0)}$ to form $\widetilde H_t^{(0)}$ are processed by $\bm\theta_{\mathrm{attn}}$ with $L$ stacked layers:
\begin{equation}
\label{eq:miloop_shallow_decoder}
\widetilde H_t^{(L)}
=
\operatorname{AttnLayers}_{\bm\theta_{\mathrm{attn}}}^{(L)}
\left(\widetilde H_t^{(0)}\right).
\end{equation}
Each layer comprises an $\operatorname{Attention}(\cdot)$ sublayer and an $\operatorname{FFN}(\cdot)$ sublayer, both equipped with residual connections~\citep{he2016skip} and RMS normalization~\citep{zhang2019rmsnorm}, following the structure in Equation~\ref{eq:attention_layer}.

The $\bm\theta_{\mathrm{fusion}}$ reuses previously computed embeddings to provide context-informed initial embeddings for the current subproblem. During rollout-based RL training, this mechanism exploits the sequential construction process to preserve structural context across steps, allowing the model to refine existing embeddings rather than reconstruct them solely from the current input, thereby reducing the difficulty of policy optimization.

\subsection{Selective Memory Update}
\label{sec:selective_memory_updates}
Memory fusion determines which historical representations are provided to the decoder, whereas selective memory update controls how the model output revises the retained memory for subsequent steps. MiLoop aligns memory updates with the generated solution, enabling earlier-computed representations to inform subsequent computation. 

\paragraph{Adaptive Gate Generation}
To adapt memory to the evolving construction state, MiLoop computes a fine-grained adaptive gate for each node and feature channel based on the retained memory $M_t^{(\mathcal{G})}$ and the newly generated candidate representation $\widetilde H_t^{(L)}$:
\begin{equation}
\label{eq:miloop_memory_gate}
\bm g_{t,i}
=
\sigma\left(
\left[\mathbf{\mathbf m}_{t,i}\Vert\widetilde{\mathbf{h}}_{t,i}^{(L)}\right]W_g^{r_{t,i}}
+
\bm b_g^{r_{t,i}}
\right),
\end{equation}
where $\mathbf m_{t,i}$ is the memory for node $i$, $\widetilde{\mathbf h}_{t,i}^{(L)}$ is its related current representation in $\widetilde H_t^{(L)}$, and $\sigma(\cdot)$ denotes the sigmoid function. Here, $r_{t,i}\in\{\mathrm{first},\mathrm{last},\mathrm{unvisited}\}$ denotes the current role of node $i$, and role-specific parameters $W_g^{r_{t,i}}\in\mathbb R^{2d\times d}$ and $\bm b_g^{r_{t,i}}\in\mathbb R^{1\times d}$ are the corresponding learnable gate parameters. These parameters are shared across nodes and construction steps within each role.

\paragraph{Memory Update Along A Rollout}
The updated memory $\widetilde{\mathbf{m}}_{t,i}$ combines the historical memory and the current candidate representation channel-wise:
\begin{equation}
\label{eq:miloop_memory_update}
\widetilde{\mathbf{m}}_{t,i}
=
(\bm 1-\bm g_{t,i})\odot\mathbf{\mathbf m}_{t,i}
+
\bm g_{t,i}\odot\widetilde{\mathbf{h}}_{t,i}^{(L)},
\end{equation}
where $\odot$ denotes element-wise multiplication. Larger gate values give greater weight to the candidate representation, whereas smaller values retain more historical information. 

The $\bm\theta_{\mathrm{attn}}$ produces a candidate representation, while $\bm\theta_{\mathrm{update}}$ controls how much each feature channel is updated. This provides a direct path for retaining memories, with the degree of retention together determined by previous and current embeddings, rather than by fixed coefficients or by fully replacing memory at each step.

\subsection{Memory-Guided Action Selection}
\label{sec:memory_guided_selection}
The updated memories are used not only for subsequent steps, but also directly for current action selection. For action selection, we use a compatibility module, following ~\citep{kool2019attention,bello2016neural}. Let $\widetilde{\mathbf{m}}_{\pi_1}$ and $\widetilde{\mathbf{m}}_{\pi_{t-1}}$ denote the updated memory states of the starting node and current node, respectively. The action score for the set of feasible actions $\mathcal N_t\subseteq\mathcal A_t$ can be computed as
\begin{equation}
\label{eq:miloop_action_scores}
\begin{aligned}
\hat{\mathbf{h}}_{(C)}^{t}
&=
\left[\widetilde{\mathbf{m}}_{\pi_1}\Vert\widetilde{\mathbf{m}}_{\pi_{t-1}}\right]W_{(C)}
+
\bm b_{(C)},\\
u_{\bm\theta}(i,\mathcal H_t)
&=
\zeta \cdot \tanh\left(
\frac{\hat{\mathbf{h}}_{(C)}^{t}\widetilde{\mathbf{m}}_{t,i}^{\mathbf{T}}}{\sqrt d}
\right),
\qquad i\in\mathcal N_t,
\end{aligned}
\end{equation}
where $W_{(C)}\in\mathbb R^{2d\times d}$ and $\bm b_{(C)}\in\mathbb R^{1\times d}$ are learnable parameters, and $\zeta$ is the clipping parameter. The scores are obtained over the feasible action set according to Equation~\ref{eq:masked_policy}. After selecting action $\pi_t$, the updated memory matrix is carried forward by node identity for reuse in subsequent construction steps, i.e., $M_t^{(\mathcal{G})}=\widetilde{M}_t^{(\mathcal{G})}$. This feedback mechanism propagates information across construction steps via an evolving memory along the policy's construction trajectory, while allowing each step to adapt to the current input.

\subsection{Training}
\label{sec:miloop_training}
MiLoop is trained using a reference-free variant of direct preference optimization (DPO)~\citep{rafailov2023dpo}, which uses objective values of complete solutions to generate a preference pair. For each instance $\mathcal G$, we sample two complete action sequences from the current policy and order them as $(\bm\pi^+,\bm\pi^-)$ so that their solutions satisfy $f(\bm y^+\mid\mathcal G)\leq f(\bm y^-\mid\mathcal G)$. Each sequence maintains memory during its construction, so action probabilities depend on both current inputs and memories formed at earlier steps. The loss function can be expressed as
\begin{equation}
\label{eq:miloop_sequence_logprob}
\log p_{\bm\theta}(\bm\pi\mid\mathcal G)
=
\sum_{t=1}^{T}
\log p_{\bm\theta}
\left(
\pi_t\mid\mathcal G,\pi_{1:t-1},s_t
\right),
\end{equation}
\begin{equation}
\label{eq:miloop_dpo_loss}
\mathcal L(\bm\theta)
\approx
-\frac{1}{B}\sum_{b=1}^{B}
\log\sigma\left(
\beta\left[
\log p_{\bm\theta}(\bm\pi_b^+\mid\mathcal G_b)
-
\log p_{\bm\theta}(\bm\pi_b^-\mid\mathcal G_b)
\right]
\right),
\end{equation}
where $\beta>0$ scales the preference logits. This objective favors the lower-cost sequence based on its relative log-probability, and the current policy generates new preference pairs throughout training, jointly optimizing input embedding, memory fusion, attention layers, gated updating, and output scoring through trajectory-level comparisons without externally labeled solutions. The detailed training process is provided in Appendix \ref{append:dpo}. We also evaluate a MiLoop variant trained with the traditional REINFORCE algorithm~\citep{williams1992reinforce}, with results reported in Table~\ref{tab:ablation_mainpaper}.

\section{Experiments}
In this section, we evaluate MiLoop against classical and neural solvers on four well-studied COPs: the Traveling Salesman Problem (TSP), the Capacitated Vehicle Routing Problem (CVRP), the Pickup-and-Delivery TSP (PDTSP), and the Knapsack Problem (KP). All experiments are run on a platform with NVIDIA GeForce RTX 4090 GPUs and an Intel Xeon Gold 6240R CPU (2.40 GHz).

\begin{table}[t]
  \centering
  \caption{Comparison of TSP and CVRP instances with uniform distribution (100 $\leq N\leq$ 10K).}
  \setlength{\tabcolsep}{4pt}
  \renewcommand{\arraystretch}{1.08}
  \resizebox{0.98\linewidth}{!}{%
  \begin{threeparttable}
    \begin{tabular}{l|ccc|ccc|ccc|ccc}
    \toprule[0.5mm]
    \multirow{2}[2]{*}{Method}
      & \multicolumn{3}{c|}{TSP100}
      & \multicolumn{3}{c|}{TSP1K}
      & \multicolumn{3}{c|}{TSP5K}
      & \multicolumn{3}{c}{TSP10K} \\
      & Obj. & Gap & Time
      & Obj. & Gap & Time
      & Obj. & Gap & Time
      & Obj. & Gap & Time \\
    \midrule
    LKH3 & 7.76 & 0.00\% & 0.34s & 23.12 & 0.00\% & 1.7m & 50.97 & 0.00\% & 12m & 71.78 & 0.00\% & 33m \\
    Concorde & 7.76 & 0.00\% & 0.36s & 23.12 & 0.00\% & 1m & 50.95 & -0.05\% & 31m & 72.00 & 0.15\% & 1.4h \\
    \midrule
    H-TSP & $-$ & $-$ & $-$ & 24.66 & 6.66\% & 48s & 55.16 & 8.21\% & 1.2m & 77.75 & 8.38\% & 2.2m \\
    UDC-$x_{50} (\alpha=50)$ & 7.78 & 0.32\% & 0.49s & 23.53 & 1.78\% & 2.92s & 54.25 & 6.43\% & 0.77s & \multicolumn{3}{c}{OOM} \\
    GLOP & 7.76 & 0.05\% & 0.5s & 23.78 & 2.85\% & 10.2s & 53.15 & 4.26\% & 1.0m & 75.04 & 4.39\% & 1.9m \\
    \midrule
    Omni-VRP aug $\times$ 8 & 7.86 & 1.29\% & 0.005s & 27.46 & 18.79\% & 4.02s & \multicolumn{3}{c|}{OOM} & \multicolumn{3}{c}{OOM} \\
    ELG aug $\times$ 8 & 7.78 & 0.22\% & 0.17s & 25.57 & 10.62\% & 4.81s & \multicolumn{3}{c|}{OOM} & \multicolumn{3}{c}{OOM} \\
    ICAM$\dagger$ aug $\times$ 8 & 7.77 & 0.15\% & 0.06s & 25.83 & 11.73\% & 1.17s & 59.98 & 17.68\% & 2.8m & \multicolumn{3}{c}{OOM} \\
    \midrule
    L2C-Insert greedy & 7.79 & 0.46\% & 0.005s & 24.22 & 4.75\% & 0.16s & 55.21 & 8.32\% & 1.94s & 77.34 & 7.75\% & 3.98s \\
    INViT-3V greedy & 7.83 & 0.95\% & 0.03s & 24.66 & 6.66\% & 9.0s & 54.49 & 6.90\% & 1.2m & 76.85 & 7.07\% & 3.7m \\
    LEHD greedy & 7.80 & 0.57\% & 0.02s & 23.84 & 3.11\% & 0.8s & 58.85 & 15.46\% & 1.5m & 91.33 & 27.24\% & 11.7m \\
    BQ greedy & 7.78 & 0.35\% & 0.04s & 23.65 & 2.30\% & 0.9s & 58.27 & 14.31\% & 22.5s & 89.73 & 25.02\% & 1.0m \\
    SIGD greedy & 7.8 & 0.42\% & 0.25s & 23.57 & 1.96\% & 1.2s & 57.19 & 12.20\% & 1.8m & 93.80 & 30.68\% & 15.5m \\
    L2R greedy & 7.99 & 2.94\% & 0.46s & 24.14 & 4.42\% & 0.05s & 53.33 & 4.62\% & 1.6s & 75.16 & 4.71\% & 3.5s \\
    Rec-NCO greedy & 7.79 & 0.33\% & 0.002s & 23.52 & 1.74\% & 0.36s & 63.59 & 24.75\% & 27.58s & 147.96 & 106.12\% & 40.32s \\
    \midrule
    MiLoop greedy & 7.82 & 0.84\% & 0.004s & 23.56 & 1.90\% & 0.89s & 52.18 & 2.38\% & 1.39s & 73.62 & 2.56\% & 2.96s \\
    MiLoop ($M$ = 100)
      & \greybg{7.76} & \greybg{0.02\%} & 0.39s
      & \greybg{23.36} & \greybg{1.05\%} & 1.27m
      & \greybg{51.82} & \greybg{1.56\%} & 1.86m
      & \greybg{73.08} & \greybg{1.81\%} & 2.34m \\
    \midrule
    \midrule
    \multirow{2}[2]{*}{Method}
      & \multicolumn{3}{c|}{CVRP100}
      & \multicolumn{3}{c|}{CVRP1K}
      & \multicolumn{3}{c|}{CVRP5K}
      & \multicolumn{3}{c}{CVRP10K} \\
      & Obj. & Gap & Time
      & Obj. & Gap & Time
      & Obj. & Gap & Time
      & Obj. & Gap & Time \\
    \midrule
    HGS & 15.56 & 0.00\% & 1.67m & 119.54 & 0.00\% & 16.67m & 590.34 & 0.00\% & 1.39h & 1078.74 & 0.00\% & 2h \\
    \midrule
    UDC-$x_{50} (\alpha=50)$ & 16.29 & 4.69\% & 0.96s & 123.00 & 2.94\% & 5.61s & 608.09 & 3.00\% & 21.04s & \multicolumn{3}{c}{OOM} \\
    GLOP-G (LKH3) & $-$ & $-$ & $-$ & 135.31 & 13.59\% & 0.41s & 639.27 & 8.28\% & 1.77s & 1146.43 & 6.27\% & 4.00s \\
    \midrule
    Omni-VRP aug $\times$ 8 & 15.98 & 2.74\% & 0.006s & 132.93 & 11.38\% & 0.72s & \multicolumn{3}{c}{OOM} & \multicolumn{3}{c}{OOM} \\
    ELG aug $\times$ 8 & 15.83 & 1.73\% & 0.31s & 125.44 & 5.12\% & 7.11s & \multicolumn{3}{c|}{OOM} & \multicolumn{3}{c}{OOM} \\
    ICAM$\dagger$  aug $\times$ 8 & 15.88 & 2.05\% & 0.09s & 125.61 & 5.19\% & 1.63s & 614.24 & 4.04\% & 3.69m & \multicolumn{3}{c}{OOM} \\
    \midrule
    L2C-Insert greedy & 16.16 & 3.89\% & 0.008s & 123.00 & 2.91\% & 7.82s & 620.76 & 5.15\% & 13.48s & 1153.79 & 6.95\% & 1.34m \\
    INViT-3V greedy & 16.93 & 8.84\% & 0.004s & 130.44 & 9.29\% & 11.23s & 636.67 & 7.84\% & 1.35m & 1162.27 & 7.74\% & 3.67m \\
    LEHD greedy & 16.21 & 4.20\% & 0.003s & 122.42 & 2.49\% & 3.58s & 636.74 & 7.85\% & 1.14m & 1199.43 & 11.18\% & 7.78m \\
    BQ greedy & 16.07 & 3.29\% & 0.01s & 121.97 & 2.16\% & 7.35s & 620.34 & 5.08\% & 5.8m & 1179.56 & 9.34\% & 8.89m \\
    SIGD greedy & 16.15 & 3.81\% & 0.035s & 125.76 & 5.31\% & 3.93s & 633.3 & 7.27\% & 21.23s & 1162.43 & 7.75\% & 48.9s \\
    L2R greedy & 16.28 & 4.68\% & 0.001s & 127.48 & 6.84\% & 5.84s & 610.68 & 3.45\% & 29.78s & 1104.44 & 2.38\% & 43.68s \\
    Rec-NCO greedy & 16.05 & 3.18\% & 0.003s & 138.65 & 16.19\% & 2.99s & 1128.68 & 91.18\% & 29.24s & 2597.54 & 140.86\% & 2.52m \\
    \midrule
    MiLoop greedy & 16.01 & 2.87\% & 0.004s & 122.26 & 2.27\% & 0.17s & 602.25 & 2.02\% & 9.18s & 1090.52 & 1.09\% & 27.25s \\
    MiLoop ($M$ = 100)
      & \greybg{15.75} & \greybg{1.21\%} & 0.23s
      & \greybg{121.34} & \greybg{1.50\%} & 46.86s
      & \greybg{598.72} & \greybg{1.42\%} & 20.99m 
      & \greybg{1086.63} & \greybg{0.73\%} & 35.82m \\
    \bottomrule[0.5mm]
    \end{tabular}%
    \begin{tablenotes}
\item[$\dagger$] ICAM is retrained only on $N=100$ instances without vary-scale training for a fair comparison.
\end{tablenotes}
\end{threeparttable}
}
  \label{tab:uniform_tsp_cvrp}
  \vspace{-10pt}
\end{table}

\subsection{Experimental Setup}
\paragraph{Problem Settings} 
We adopt established synthetic data generation protocols for TSP, CVRP, and KP~\citep{kool2019attention,kwon2020pomo} and PDTSP~\citep{li2021pdtsp}. For TSP and CVRP, we use test sets with uniformly distributed nodes at four sizes: 100, 1K, 5K, and 10K. The capacity is fixed to $50$ for CVRP instances. For PDTSP, we use the existing 100-node test set~\citep{li2021pdtsp} and generate 128 instances at each of three larger sizes: 200, 500, and 1K nodes. For KP, we follow the evaluation protocol of BQ~\citep{drakulic2023bq} to assess solution quality across multiple sizes ($ 200, 500, 1000$) and capacities. To assess cross-distribution generalization, we evaluate MiLoop on four benchmarks: (1) symmetric \texttt{EUC\_2D} instances from TSPLib~\citep{reinelt1991tsplib}; (2) CVRPLib Set-X~\citep{uchoa2017cvrplib_setx}; (3) CVRPLib Set-AGS~\citep{arnold2019cvrplib_xxl}; and (4) instances from the 8th DIMACS Implementation Challenge~\citep{dimacs}.

\paragraph{Model \& Training Settings}
MiLoop uses three attention layers with structural perception~\citep{zhou2026icam}, embedding dimension $d=128$, feed-forward hidden dimension $512$, logit clipping $\zeta=10$~\citep{kool2019attention}, and preference scaling $\beta=0.1$. Training uses uniformly distributed $100$-node instances for three VRPs, and $200$-item instances for KP~\citep{kwon2020pomo}. Models are trained for $110$ epochs ($20$ for KP), with $2{,}500$ batches per epoch. Due to memory constraints, we use batch sizes of $120$, $100$, $120$, and $40$ for TSP, CVRP, PDTSP, and KP, respectively. We use AdamW~\citep{loshchilov2017adamw} with an initial learning rate of $10^{-4}$, which is decayed by a factor of 0.1 at epoch $101$ ($15$ for KP). Further details are provided in Appendix~\ref{append:model_setting}.

\paragraph{Baselines}
We compare MiLoop with following methods: (1) \textbf{Classical Solver}: Concorde~\citep{applegate2006concorde}, LKH3~\citep{LKH3}, HGS~\citep{HGS}; (2) \textbf{Constructive NCO with Static Embeddings}: POMO~\citep{kwon2020pomo}, Omni-VRP~\citep{zhou2023omni}, and ELG~\citep{gao2023elg}, ICAM~\citep{zhou2026icam}, Heter-AM~\citep{li2021pdtsp}, and URS~\citep{zhou2026urs}; (3) \textbf{Constructive NCO with Dynamic Embeddings}: BQ~\citep{drakulic2023bq}, LEHD~\citep{luo2023lehd}, SIGD~\citep{pirnay2024self}, L2C-Insert~\citep{luo2025insertion}, INViT~\citep{fang2024invit}, L2R~\citep{zhou2025L2R} and Rec-NCO~\citep{dernedde2026recurrent}; (4) \textbf{Two-stage NCO}: H-TSP~\citep{pan2023htsp}, GLOP~\citep{ye2023glop}, and UDC~\citep{zheng2024udc}.

\paragraph{Metrics and Inference}
We report average objective values (Obj.), optimality gaps (Gap), and average times (Time). ``OOM'' indicates running out of memory. MiLoop uses greedy and multi-start decoding ($M=100$) from distinct starting nodes~\citep{kwon2020pomo}. For $N>1{,}000$, we limit the candidate set to $200$ nodes per step for efficiency~\citep{chen2025projection} ($400$ for CVRP to account for capacity constraints in addition to geometric proximity~\citep{zhou2025L2R}).

\begin{table}[t]
  \centering
  \caption{Comparison on TSPLib ($0 \textless N \leq 85,900$) and CVRPLib instances ($0 \textless N \leq 30,000$).}
    \resizebox{0.98\textwidth}{!}{
    \begin{threeparttable}
        \begin{tabular}{l|c|c|c|c|c|c|c|c|c|c}
    \toprule[0.5mm]
    \multirow{2}[4]{*}{Method} & \multicolumn{5}{c|}{TSPLIB}           & \multicolumn{5}{c}{CVRPLIB} \\
\cmidrule{2-11}          & (0,1K] & (1K, 5K]  & (5K, 100K) & All   & Solved \# & (0,1K] & (1K, 7K] & (7K,30K] & All   & Solved \# \\
    \midrule
    UDC α=50 & 4.28\% & 11.89\% & 24.81\%  & 7.13\% $\dagger$ & 73/81 & 7.12\% & 14.06\%  & OOM   & 7.33\% $\dagger$ & 103/110 \\
    GLOP  & 5.55\% & 12.92\% & 11.99\% & 8.35\% & 81/81 & 18.86\% & 19.89\% & 24.42\% & 19.20\% & 110/110 \\
    \midrule
    Omni-VRP aug$\times$8& 13.04\% & 39.14\% & 79.65\%  & 26.38\% $\dagger$ & 78/81 & 11.89\% & 165.02\% & 223.03\%  & 21.65\%  $\dagger$ & 106/110 \\
    ELG aug$\times$8 & 3.13\% & 11.13\% & OOM   & 5.53\% $\dagger$ & 70/81 & 6.03\% & 16.37\%  & OOM   & 6.23\% $\dagger$ & 102/110 \\
    ICAM aug$\times$8 & 3.89\% & 14.23\% & 17.52\%  & 7.52\% $\dagger$ & 74/81 & 6.21\% & 16.86\% & OOM   & 6.62\% $\dagger$ & 104/110 \\
    \midrule
    L2C-Insert greedy & 3.27\% & 11.26\% & 24.56\% & 8.07\% & 81/81 & 6.91\% & 14.75\% & 25.00\% & 8.18\% & 110/110 \\
    INViT-3V greedy & 8.30\% & 10.84\% & 9.55\% & 9.15\% & 81/81 & 13.51\% & 20.77\% & 26.53\% & 14.49\% & 110/110 \\
    LEHD greedy & 2.71\% & 10.79\% & 39.02\%  & 8.25\%  $\dagger$& 78/81 & 12.54\% & 21.48\% & 33.71\%  & 13.27\% $\dagger$ & 106/110 \\
    BQ greedy & 3.20\% & 9.68\% & 29.79\%  & 6.52\%  $\dagger$ & 75/81 & 8.96\% & 27.97\% & 53.27\%  & 10.51\% $\dagger$ & 106/110 \\
    SIGD greedy & 2.79\% & 10.65\% & 117.40\%  & 16.58\% $\dagger$ & 79/81 & 19.19\% & 18.20\% & 48.91\% & 20.78\% & 110/110 \\
    L2R greedy  & 6.21\% & 8.75\% & 7.34\% & 7.04\% & 81/81 & 8.18\% & 11.95\% & 11.08\% & 8.48\% & 110/110 \\
    Rec-NCO  greedy& 3.08\% & 62.98\% & 776.86\%  & 106.61\% $\dagger$ & 80/81 & 15.64\% & 210.23\% & 1093.44\% & 81.50\%  & 110/110 \\
    \midrule
    MiLoop greedy& 4.85\% & 6.39\% & 6.50\% & 5.48\% & 81/81 & 5.81\% & 9.55\% & 10.72\% & 6.21\% & 110/110 \\
    MiLoop ($M$ = 100) & \greybg{1.31\%} & \greybg{3.17\%} & \greybg{3.16\%} & \greybg{2.04\%} & 81/81 & \greybg{4.15\%} & \greybg{7.98\%} & \greybg{8.80\%} & \greybg{4.54\%} & 110/110 \\
    \bottomrule[0.5mm]
    \end{tabular}%
\begin{tablenotes}
\item[\textdagger] Some instances are skipped due to the OOM issue.
\end{tablenotes}
\end{threeparttable}
}
  \label{tab:benchmark}%
  \vspace{-10pt}
\end{table}%

\subsection{Performance Evaluation}
\paragraph{Results on Synthetic Datasets}
Results are reported in~\cref{tab:uniform_tsp_cvrp} for TSP and CVRP ($100$--$10\mathrm{K}$ nodes), and in~\cref{tab:uniform_pdtsp,tab:uniform_kp} in Appendix~\ref{append:results_uniform_pdtsp_kp} for PDTSP ($100$--$1\mathrm{K}$ nodes) and KP ($200$--$1\mathrm{K}$ items), respectively. For TSP and CVRP, MiLoop ($M=100$) outperforms all neural baselines across all settings. Notably, MiLoop substantially outperforms the state-of-the-art RL-based L2R with dynamic embeddings on large-scale instances and achieves a gap of $0.02\%$ on TSP100. On PDTSP, MiLoop also achieves the smallest gap among the neural baselines across all tested scales. Additionally, MiLoop also achieves the lowest gaps among neural solvers in $11$ of $12$ KP settings, demonstrating cross-scale generalization beyond routing. Overall, these results demonstrate the strong cross-scale generalization of MiLoop with selective memory propagation.

\paragraph{Results on Benchmark Datasets}
\begin{wraptable}[13]{r}{0.4\textwidth}
\vspace{-13pt}
\centering
\caption{Comparison on ultra-large-scale TSP instances (greedy decoding).}
\label{tab:dimacs_tsp}
\resizebox{0.4\textwidth}{!}{
\begin{tabular}{l|cccc}
\toprule[0.5mm]
Instance & LEHD & INViT & L2R & MiLoop \\
\midrule
E10k.0  & 24.63\% & 6.64\% & 4.56\% & \greybg{1.90\%} \\
E10k.1  & 26.51\% & 7.08\% & 4.59\% & \greybg{2.28\%} \\
E10k.2  & 24.74\% & 7.38\% & 4.77\% & \greybg{2.92\%} \\
E31k.0  & OOM     & 6.97\% & 4.82\% & \greybg{4.18\%} \\
E31k.1  & OOM     & 7.22\% & 4.74\% & \greybg{2.42\%} \\
E100k.0 & OOM     & OOM    & 4.71\% & \greybg{2.52\%} \\
E100k.1 & OOM     & OOM    & 4.98\% & \greybg{2.55\%} \\
E316k.0 & OOM     & OOM    & 4.82\% & \greybg{2.86\%} \\
E1M.0   & OOM     & OOM    & 5.02\% & \greybg{2.97\%} \\
E3M.0   & OOM     & OOM    & 5.05\% & \greybg{2.64\%} \\
E10M.0  & OOM     & OOM    & 5.05\% & \greybg{2.75\%} \\
\bottomrule[0.5mm]
\end{tabular}
}
\end{wraptable}
As shown in~\cref{tab:benchmark}, MiLoop successfully solves all $81$ TSPLib and $110$ CVRPLib instances. With $M=100$, it achieves the lowest reported average gaps across all scale groups. We also examine scalability on substantially larger TSP instances. As shown in~\cref{tab:dimacs_tsp}, MiLoop still achieves the highest solution quality among comparable neural methods on instances with up to \textbf{10 million} nodes. Notably, as a purely learning-based constructive neural solver, MiLoop is trained only on uniformly distributed $100$-node instances, yet consistently produces high-quality solutions for unprecedented problem instances with up to 10 million nodes. These results demonstrate our strong generalization across both problem scales and data distributions.

\subsection{Discussion}
\paragraph{Comparison with Existing Recurrent Embeddings}

\begin{wraptable}[10]{r}{0.4\textwidth}
  \centering
  \vspace{-11pt}
  \caption{Comparison of different recurrent embeddings.}
  \label{tab:recurrent_compare}
  \setlength{\tabcolsep}{4pt}
  \renewcommand{\arraystretch}{1.08}
  \resizebox{\linewidth}{!}{%
    \begin{tabular}{cc|c|cc}
      \toprule[0.5mm]
      \multicolumn{2}{c|}{Input Fusion}
        & \multirow{2}{*}{\shortstack{Gated\\Update}}
        & TSP100 & TSP1K \\
      \cmidrule(lr){1-2}
      RSE & MiLoop
        & & Gap & Gap \\
      \midrule
      $\checkmark$ & $\times$ & $\times$
        & 0.88\% & 9.17\% \\
      $\times$ & $\checkmark$ & $\times$
        & 1.17\% & 2.90\% \\
      $\times$ & $\checkmark$ & $\checkmark$
        & \greybg{\textbf{0.84\%}}
        & \greybg{\textbf{1.90\%}} \\
      \bottomrule[0.5mm]
    \end{tabular}%
  }
\end{wraptable}
Rec-NCO~\citep{dernedde2026recurrent} is most closely related to MiLoop, as both reuse embeddings across steps. Rec-NCO uses non-linear recurrent state embeddings (RSE) with residual connections as inputs and employs a two-stage supervised learning approach: pretraining a base model, followed by training a separate recurrent module. In contrast, MiLoop jointly learns its shallow policy and memory through complete RL rollouts without external solution labels. It combines current--history fusion with post-attention gates to selectively retain and update memory across nodes and feature channels. We compare full-RSE and fusion-only variants, both without gated updates, under otherwise identical settings. Table~\ref{tab:recurrent_compare} highlights the importance of \textbf{current information fusion and selective memory updates} for cross-scale generalization beyond recurrence alone.

\begin{figure}[t]
    \centering
    \includegraphics[width=0.98\linewidth]{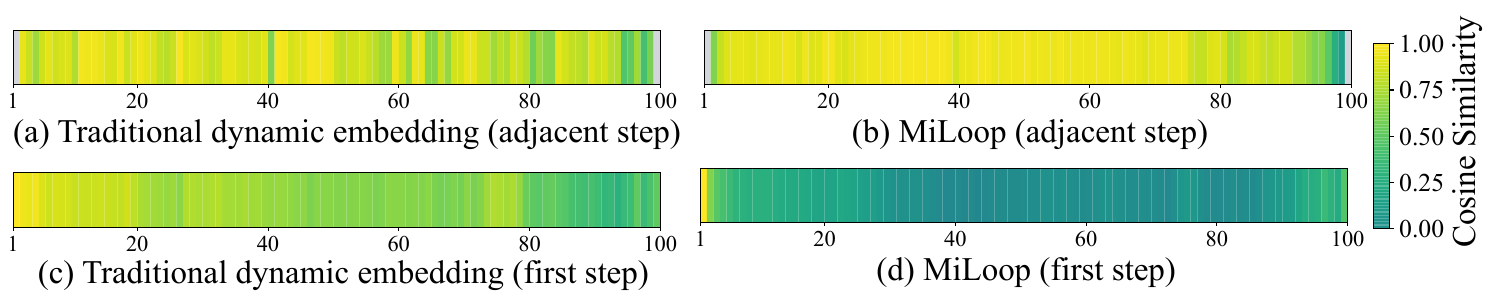}
    \caption{Cosine similarity of node embeddings during TSP construction for a traditional dynamic-embedding model variant and MiLoop. We use the adjacent step or the first step as the reference.}
    \label{fig:memory_propagation}
    \vspace{-5pt}
\end{figure}

\paragraph{Stepwise Memory Propagation}
We compare MiLoop on TSP against a trained three-layer dynamic-embedding baseline that recomputes node embeddings for the current subproblem at each construction step. As shown in~\cref{fig:memory_propagation}, the baseline exhibits greater adjacent-step fluctuations while remaining relatively similar to its first-step representations. In contrast, MiLoop maintains strong representational continuity between adjacent steps while forming representations distinct from the first step after multiple steps. Such local continuity and cumulative changes provide intuitive support for MiLoop's stepwise memory update mechanism. The detailed experimental results are provided in \cref{tab:ablation_mainpaper}. See Appendix \ref{append:memory_effectiveness} for more detailed discussions. 

\subsection{Ablation Study}
We conduct an ablation study on TSP to validate the effect of memory fusion, gated updates, attention depth, and the training loss.
As shown in \cref{tab:ablation_mainpaper}, removing either memory component increases the gaps across all tested scales, supporting their complementary roles in incorporating current information and selectively retaining historical representations.
Increasing the attention depth from three to six layers improves the model without memory propagation, but it still falls short of the three-layer MiLoop.
On TSP10K, MiLoop achieves a lower gap ($2.56\%$ vs.\ $4.39\%$) with approximately half the inference time, highlighting the benefit of stepwise memory reuse over simply increasing per-step depth.
MiLoop with REINFORCE still exhibits strong cross-scale robustness, while replacing the loss with DPO yields a further overall improvement.

\begin{table}[t]
\centering
\caption{Ablation study of MiLoop on TSP. RF denotes REINFORCE algorithm~\citep{williams1992reinforce}. }
\label{tab:ablation_mainpaper}
\resizebox{0.98\textwidth}{!}{
\begin{tabular}{ccc|c|c|cc|cc|cc}
\toprule[0.5mm]
\multicolumn{2}{c}{Memory}
& Looped
& Attention
& \multirow{2}{*}{Train}
& \multicolumn{2}{c|}{TSP100}
& \multicolumn{2}{c|}{TSP1K}
& \multicolumn{2}{c}{TSP10K} \\
Fusion & Update & Embedding & Layers &
& Obj. (Gap) & Time
& Obj. (Gap) & Time
& Obj. (Gap) & Time \\
\midrule

$\checkmark$ & $\times$ & $\checkmark$ & 3 & DPO
& 7.85 (1.17\%) & 0.004s
& 23.79 (2.90\%) & 0.9s
& 74.24 (3.43\%) & 2.96s \\

$\times$ & $\checkmark$ & $\checkmark$ & 3 & DPO
& 7.90 (1.79\%) & 0.004s
& 24.95 (7.93\%) & 0.9s
& 78.86 (9.98\%) & 2.96s \\

$\times$ & $\times$ & $\times$ & 3 & DPO
& 7.88 (1.61\%) & 0.004s
& 24.64 (6.59\%) & 0.9s
& 75.82 (5.63\%) & 2.96s \\

$\times$ & $\times$ & $\times$ & 6 & DPO
& 7.85 (1.18\%) & 0.008s
& 24.45 (5.78\%) & 1.72s
& 74.93 (4.39\%) & 5.84s \\

\midrule
$\checkmark$ & $\checkmark$ & $\checkmark$ & 3 & RF
& 7.86 (1.26\%) & 0.004s
& 24.23 (4.81\%) & 0.89s
& 74.85 (4.28\%) & 2.96s \\

$\checkmark$ & $\checkmark$ & $\checkmark$ & 3 & DPO
& \greybg{7.82 (0.84\%)} & 0.004s
& \greybg{23.56 (1.90\%)} & 0.9s
& \greybg{73.62 (2.56\%)} & 2.96s \\

\bottomrule[0.5mm]
\end{tabular}
\vspace{-10pt}
}
\end{table}

\section{Conclusion}
In this work, we proposed MiLoop, an RL-based framework that leverages the multi-step computation already required by construction rollouts for selective memory propagation. Each rollout provides solution-quality feedback while propagating structural representations, enabling a three-layer policy to learn effective dynamic embeddings without external solution labels or training-time search-space pruning. MiLoop fuses current embeddings with historical memory before the attention layers and applies adaptive gated updates afterward, allowing the updated representations to support both current decisions and stepwise reuse. Extensive experiments across four COPs, including instances with up to 10 million nodes, demonstrate high solution quality and strong cross-size generalization. Ablation studies further support the complementary roles of current-information fusion and gated memory updates. Together, these results highlight selective memory reuse as an effective way to reduce reliance on deep per-step recomputation in RL-based constructive NCO. In the future, we will explore more effective memory fusion and update mechanisms to further improve progressive representation reuse. We also plan to extend MiLoop to multi-task learning, enabling a single policy to solve different types of combinatorial optimization problems.

\subsection*{AI use statement}

In this work, we used generative AI tools solely for
translation, language editing, and proofreading. These tools helped
improve clarity and fluency and correct grammatical, spelling, and
punctuation errors. We reviewed all AI-assisted translations and edits
to ensure that they accurately conveyed the intended meaning and did
not alter the technical content or scientific claims. All core research
ideas, study design, data analysis, and conclusions were independently
developed by the authors. We take full responsibility for the final
content of this work, including text produced with the assistance of
generative AI.

\bibliography{references}
\bibliographystyle{iclr2027_conference}

\clearpage
\appendix
\startcontents[appendix]
\printcontents[appendix]{}{1}{%
    \section*{Appendix Contents}%
    \setcounter{tocdepth}{3}%
}
\clearpage

\section{Related Work}
\label{append:related_work}
\subsection{Constructive NCO with Static Embeddings}
The constructive NCO solvers with static embeddings encode the instance prior to solution construction and reuse the resulting node embeddings throughout the autoregressive construction. This paradigm can be traced back to the recurrent node encoding used in Pointer Networks~\citep{vinyals2015pointer} and was later adopted in a reinforcement learning framework~\citep{bello2016neural}. Moving beyond this, ~\citep{kool2019attention, kwon2020pomo} adopts self-attention to capture global node relationships. Building on these static representations, subsequent efforts improve generalization by incorporating solution symmetry and structural biases, making learned policies more robust across different instances. ~\citep{kim2022symnco,gao2023elg,zhou2026icam, kwon2021matnet}. Beyond these improvements, the same encoding paradigm has been utilized for multi-task learning, broadening its applicability across various problem variants~\citep{zhou2023omni,yi2026radar,li2021pdtsp,huang2025reld,zhou2026urs,chen2026space}. Complementing the use of node embeddings for autoregressive decisions, heatmap-based methods instead encode graph information into edge-selection scores that guide subsequent solution construction and search~\citep{qiu2022dimes,sun2023difusco,li2023T2T,li2024fastT2T}. However, fixed node embeddings do not explicitly adapt to the evolving subproblem during construction, potentially limiting generalization. This limitation motivates the development of methods based on dynamic embeddings.

\subsection{Constructive NCO with Dynamic Embeddings}

Dynamic-embedding methods repeatedly encode the subproblem during solution construction and use the updated node representations to guide action selection. Early methods update embeddings through several encoder layers~\citep{xin2020stepwise} or the final encoder layer~\citep{xin2021multidecoder}, followed by a lightweight decoder for action selection. While earlier studies focus on in-domain performance, subsequent research shifts model capacity toward a heavier decoder~\citep{luo2023lehd,drakulic2023bq,chen2025projection,luo2025insertion,luo2026efficient}, thereby improving out-of-domain performance.
However, training these models relies heavily on a large number of high-quality labels. To address this dependence, subsequent studies propose self-improvement learning (SIL), which generates solutions using the current policy and uses them as pseudo-labels for supervised training~\citep{luo2024Boosting,pirnay2024self,xiao2025dgl}. Although SIL avoids the cost of obtaining high-quality labels, the additional iterations required to generate high-quality pseudo-labels incur prohibitive computational overhead. Additionally, MTL-KD~\citep{zheng2025mtl_kd} leverages knowledge distillation to further reduce data dependency and enhance large-scale generalization. More recently, L2R~\citep{zhou2025L2R} and INViT~\citep{fang2024invit} employ reinforcement learning to train dynamic-embedding models. Specifically, they dynamically encode reduced subproblems using a multi-layer decoder, improving generalization to large-scale instances. However, training dynamic-embedding models incurs substantial computational overhead because node embeddings are recomputed at every autoregressive step. As a result, RL-based dynamic-embedding methods often train in a reduced space to ease training, thereby excluding actions required for high-quality solutions. To overcome this limitation, we propose MiLoop, which incorporates selective memory into dynamic node embeddings, enabling a parameter-efficient model to learn directly from the entire search space.

\subsection{Looped Transformer}

Looped Transformers repeatedly apply shared blocks to refine hidden representations, increasing effective computational depth without a proportional increase in independent parameters~\citep{dehghani2018universal,yang2024looped}. 
Recent work exploits this principle for latent reasoning and test-time computation~\citep{geiping2026scaling}, while Mixture-of-Recursions~\citep{bae2026mixture} and LoopFormer~\citep{jeddi2026loopformer} investigate token-adaptive recursion and budget-conditioned depth, respectively. A complementary line introduces recurrence across sequence positions or segments.
Feedback Transformer~\citep{fan2020addressing} exposes earlier high-level representations to subsequent computations, whereas Recurrent Memory Transformer~\citep{bulatov2022recurrent} and Block-Recurrent Transformers~\citep{hutchins2022block} propagate memory across segments, with the latter incorporating gated state updates. MiLoop shares the motivation of reusing computed representations, but focuses on selective memory propagation across autoregressive solution-construction steps in NCO.

\clearpage
\section{Implementation Details}
\label{append:implementation_details}
The following section introduces the technical details of MiLoop for different problems.
\subsection{Capacitated Vehicle Routing Problem}
\label{append:implementation_cvrp}

\subsubsection{Problem Setup}
A CVRP instance $\mathcal G$ consists of $N$ customer nodes and one depot node. We index the depot by $0$ and the customers by $1,\ldots,N$. Each node $i$ has coordinates $\bm x_i$, and each customer $i$ has a demand $q_i$ that must be fulfilled. A feasible solution $\bm y\in\Omega(\mathcal G)$ is a set of routes, each starting and ending at the depot. Every customer is visited exactly once, with its demand fully satisfied in that visit. The total demand of the customers on each route must not exceed the vehicle capacity $C$.

\subsubsection{MiLoop for CVRP}
\label{append:miloop_cvrp}
MiLoop for CVRP follows the memory fusion, selective update, and action selection procedure in Section~\ref{sec:methodology}, with task-specific inputs and a capacity-conditioned output module. We index the depot by $0$ and the $N$ customers by $1,\ldots, N$. Let $q_i$ be the demand of customer $i$, $C$ the vehicle capacity, and $c_t\in[0,1]$ the remaining capacity normalized by $C$. We use normalized demands $\bar q_i=q_i/C$ and set $\bar q_0=0$. Construction starts at the depot, i.e., $\pi_1=0$.

\paragraph{Initial Embedding and Feasibility}
At step $t\geq2$, the candidate set $\mathcal A_t$ includes all unvisited customers and, whenever $\pi_{t-1}\neq0$, the depot as a possible return action. The feasible action set is
\begin{equation}
\mathcal N_t=\{i\in\mathcal A_t:\bar q_i\leq c_t\}.
\label{eq:cvrp_feasible_actions}
\end{equation}

The model input consists of a depot node, a current node, and unvisited nodes. We normalize their coordinates at each step using the coordinate normalization method~\citep{fang2024invit,fu2021attgcn_mcts}. Denoting the transformed coordinates by $\widetilde{\bm x}_{t,i}$. The shared linear projection is
\begin{equation}
\mathbf h_{t,i}^{(0)}
=\left[\widetilde{\bm x}_{t,i}\Vert\bar q_i\right]W^{(0)}+\bm b^{(0)},
\qquad W^{(0)}\in\mathbb R^{3\times d}.
\label{eq:cvrp_input_embedding}
\end{equation}
Separate projections $W_{\mathrm{first}}^{(0)}$ and $W_{\mathrm{last}}^{(0)}$ distinguish the depot and current-node roles, as in the TSP implementation.

\paragraph{Stepwise Memory Fusion and Selective Update}
We maintain $M_t^{(\mathcal G)}\in\mathbb R^{(N+1)\times d}$, indexed by the original node identities, including the depot. At the first node, the current embeddings directly serve as the decoder input and initialize the memories to zero. At subsequent steps, the depot, current-node, and candidate memories are retrieved and fused with their current embeddings through the three fusion metrices in Equation~\ref{eq:miloop_input_fusion}. After the attention layers, the role-specific gates in Equations~\ref{eq:miloop_memory_gate} and~\ref{eq:miloop_memory_update} produce the updated memories. We denote the updated depot and current node by $\widetilde{\mathbf{m}}_{t}^{\mathrm{dep}}$ and $\widetilde{\mathbf m}_{t}^{\mathrm{cur}}$, and the candidate memories by $\widetilde M_{t,\mathcal A_t}$. They are used for current action scoring and written back by node indices for the next step. If the depot occurs in multiple roles, the embeddings are updated separately and written in depot, current-node, then candidate order, with later writes taking precedence. Memory persists across depot returns as a complete solution is constructed.

\paragraph{Capacity-Conditioned Context}
The updated depot and current-node memories first form a context embedding and are then modulated by the remaining capacity, as in~\citep {yu2026rethinking}.

\begin{equation}
\begin{aligned}
\hat{\mathbf h}_{(C)}^t
&=\left[\widetilde{\mathbf m}_{t}^{\mathrm{dep}}\Vert\widetilde{\mathbf m}_{t}^{\mathrm{cur}}\right]W_{(C)}+\bm b_{(C)},\\
\widetilde{\mathbf h}_{(C)}^t
&=\left(\left[\hat{\mathbf h}_{(C)}^t\Vert c_t\right]W_{\mathrm{cat}}+\bm b_{\mathrm{cat}}\right)
\odot\left(\bm 1+\bm\gamma(c_t)\right)+\bm\beta(c_t),
\end{aligned}
\label{eq:cvrp_capacity_context}
\end{equation}
where $W_{\mathrm{cat}}\in\mathbb R^{(d+1)\times d}$, and $\bm\gamma(c_t)$ and $\bm\beta(c_t)$ are linear projections that maps the scalar capacity to $d$-dimensional vectors.

\paragraph{Memory-Guided Action Selection}
An $\operatorname{Attention}(\cdot)$ further refines the capacity-conditioned query by aggregating the updated candidate memories:
\begin{equation}
\widetilde{\mathbf h}_{\mathrm{out}}^t
=\operatorname{Attention}\!\left(
\widetilde{\mathbf h}_{(C)}^t W_Q^{\mathrm{out}},
\widetilde M_{t,\mathcal A_t}W_K^{\mathrm{out}},
\widetilde M_{t,\mathcal A_t}W_V^{\mathrm{out}};
B_t^{\mathrm{out}}
\right).
\label{eq:cvrp_light_decoder}
\end{equation}
\begin{equation}
\mathbf h_{\mathrm{out}}^t = \widetilde{\mathbf h}_{(C)}^t + \widetilde{\mathbf h}_{\mathrm{out}}^t 
\end{equation}
For candidate $i$, the distance bias is $B_{t,i}^{\mathrm{out}}=-\alpha_{\mathrm{out}}\log_2|\mathcal A_t|\,\|\widetilde{\bm x}_{t,\pi_{t-1}}-\widetilde{\bm x}_{t,i}\|_2$, with a learnable coefficient initialized to $1$ and padding masked out. This aggregation uses all candidates in $\mathcal {A}_{t} $ for the whole observation space~\citep{yu2026rethinking}. The final logits are
\begin{equation}
 u_{\bm\theta}(i,\mathcal H_t)=
 \begin{cases}
  \zeta \cdot \tanh\left(\frac{\mathbf h_{\mathrm{out}}^t\widetilde{\mathbf m}_{t,i}^{\mathbf T}}{\sqrt d}\right), & i\in\mathcal N_t,\\
  -\infty, & i\in\mathcal A_t\setminus\mathcal N_t.
 \end{cases}
\label{eq:cvrp_action_scores}
\end{equation}
A softmax yields the action probabilities. Selecting customer $i$ decreases $c_t$ by $\bar q_i$, while selecting the depot resets it to $1$. Construction terminates after all customers have been served and the vehicle has returned to the depot.

\subsection{Pickup-and-Delivery TSP}
\label{append:implementation_pdtsp}

\subsubsection{Problem Setup}

A PDTSP instance $\mathcal G$ consists of $N=1+n$ nodes: one depot and $n/2$ pickup--delivery pairs. We index the depot by $0$, the pickup nodes by $1,\ldots,n/2$, and their corresponding delivery nodes by $n/2+1,\ldots,n$, so that pickup node $i$ is paired with delivery node $i+n/2$. Each node $i$ has coordinates $\bm x_i$. A feasible solution $\bm y=(\pi_1,\ldots,\pi_N)\in\Omega(\mathcal G)$ specifies a single tour that starts at the depot, visits every pickup and delivery node exactly once, and returns to the depot. Each pickup node must be visited before its paired delivery node. The objective is to minimize the total Euclidean travel distance, including the final return to the depot.

\subsubsection{MiLoop for PDTSP}
\label{append:miloop_pdtsp}
MiLoop for PDTSP follows the memory fusion, selective memory update, and action selection procedure in Section~\ref{sec:methodology}. For each pickup node, both its current input embedding and memory fusion incorporate information from its paired delivery node.

\paragraph{Current Input Embedding and Feasibility}
At construction step $t\geq2$, the node set $\mathcal A_t$ contains all unvisited pickup and delivery nodes. An unvisited pickup node is a feasible action, whereas a delivery node is feasible only after its paired pickup node has been visited. The feasible action set is
\begin{equation}
\mathcal N_t
=\{i\in\mathcal A_t:i\leq n/2\}
\cup\{i+n/2\in\mathcal A_t:i\in\{\pi_1,\ldots,\pi_{t-1}\},\ 1\leq i\leq n/2\}.
\label{eq:pdtsp_feasible_actions}
\end{equation}
The model input consists of the starting node $\pi_1=0$, the current node $\pi_{t-1}$, and the unvisited nodes in $\mathcal A_t$. Let $\bm x_{t,i}\in\mathbb R^{1\times2}$ denote the normalized coordinates of node $i$ at step $t$. We map the coordinates of the depot and delivery nodes, and the coordinates of each pickup node together with its paired delivery node, to $d$-dimensional embeddings through separate linear projections:
\begin{equation}
\begin{aligned}
\mathbf h_{t,0}^{(0)}
&=\bm x_{t,0}W_{\mathrm{depot}}^{(0)}+\bm b_{\mathrm{depot}}^{(0)},\\
\mathbf h_{t,i}^{(0)}
&=\left[\bm x_{t,i}\Vert\bm x_{t,i+n/2}\right]W_{\mathrm{pickup}}^{(0)}+\bm b_{\mathrm{pickup}}^{(0)},\\
\mathbf h_{t,i+n/2}^{(0)}
&=\bm x_{t,i+n/2}W_{\mathrm{delivery}}^{(0)}+\bm b_{\mathrm{delivery}}^{(0)},
\qquad i=1,\ldots,n/2,
\end{aligned}
\label{eq:pdtsp_input_embedding}
\end{equation}
where $W_{\mathrm{depot}}^{(0)},W_{\mathrm{delivery}}^{(0)}\in\mathbb R^{2\times d}$, $W_{\mathrm{pickup}}^{(0)}\in\mathbb R^{4\times d}$, and the corresponding biases are learnable parameters. As in the TSP implementation, separate linear projections distinguish the starting node and current node, yielding $\bar{\mathbf h}_{\pi_1}^{(0)}=\mathbf h_{t,0}^{(0)}W_{\mathrm{first}}^{(0)}$ and $\bar{\mathbf h}_{\pi_{t-1}}^{(0)}=\mathbf h_{t,\pi_{t-1}}^{(0)}W_{\mathrm{last}}^{(0)}$. These embeddings form $H_t^{(0)}\in\mathbb R^{(2+|\mathcal A_t|)\times d}$.

\paragraph{Stepwise Memory Fusion and Selective Update}
We maintain a global node-wise memory matrix $M_t^{(\mathcal G)}\in\mathbb R^{N\times d}$, including the depot. At the first call, we directly use $\widetilde H_t^{(0)}=H_t^{(0)}$ and use the corresponding current embeddings as the initial memories. In subsequent construction steps, the historical representations of the starting node, the current node, and the unvisited nodes are retrieved and fused with their current embeddings. For each pickup node, memory fusion also uses the historical representation and current embedding of its paired delivery node. The details of the fusion between pickup and delivery nodes are introduced below.

Let $r\in\{\mathrm{last},\mathrm{unvisited}\}$ denote the role of the current node or an unvisited node, respectively. The fused embeddings of pickup node $i$ and delivery node $i+n/2$, when appearing in role $r$, are
\begin{equation}
\begin{aligned}
\widetilde{\mathbf h}_{t,i}^{(0),r}
&=\left[\mathbf m_{t,i}\Vert\mathbf h_{t,i}^{(0),r}\right]W_{\mathrm{fusion}}^{r,\mathrm{pickup}}
+\left[\mathbf m_{t,i+n/2}\Vert\mathbf h_{t,i+n/2}^{(0),r}\right]W_{\mathrm{fusion}}^{r,\mathrm{delivery}},\\
\widetilde{\mathbf h}_{t,i+n/2}^{(0),r}
&=\left[\mathbf m_{t,i+n/2}\Vert\mathbf h_{t,i+n/2}^{(0),r}\right]W_{\mathrm{fusion}}^{r,\mathrm{delivery}},
\end{aligned}
\label{eq:pdtsp_input_fusion}
\end{equation}
where $W_{\mathrm{fusion}}^{r,\mathrm{pickup}},W_{\mathrm{fusion}}^{r,\mathrm{delivery}}\in\mathbb R^{2d\times d}$ are learnable fusion matrices, shared across nodes within each role. The current node and unvisited nodes use separate fusion matrices. For the current pickup node, $W_{\mathrm{last}}^{(0)}$ is applied to both its embedding and the embedding of its paired delivery node. The starting node uses $\widetilde{\mathbf h}_{\pi_1}^{(0)}=[\mathbf m_{t,0}\Vert\bar{\mathbf h}_{\pi_1}^{(0)}]W_{\mathrm{fusion}}^{\mathrm{first}}$. The fused embeddings are then processed by $L$ stacked attention layers to obtain $\widetilde H_t^{(L)}$, as in Equation~\ref{eq:miloop_shallow_decoder}.

After the attention layers, adaptive gates combine the historical memory and the current candidate representation:
\begin{align}
\bm g_{t,i}
&=\sigma\left(\left[\mathbf m_{t,i}\Vert\widetilde{\mathbf h}_{t,i}^{(L)}\right]W_g^{r_{t,i}}+\bm b_g^{r_{t,i}}\right),
\label{eq:pdtsp_memory_gate}\\
\widetilde{\mathbf m}_{t,i}
&=(1-\bm g_{t,i})\odot\mathbf m_{t,i}
+\bm g_{t,i}\odot\widetilde{\mathbf h}_{t,i}^{(L)},
\label{eq:pdtsp_memory_update}
\end{align}
where $r_{t,i}\in\{\mathrm{first},\mathrm{last},\mathrm{unvisited}\}$ denotes the current role of node $i$. The gate parameters are shared across nodes and construction steps within each role. All unvisited nodes participate in the attention layers and memory updates, including delivery nodes whose paired pickup nodes have not been visited. The updated memories support current action scoring and are carried forward by node identity for reuse in subsequent construction steps.

\paragraph{Memory-Guided Action Selection}
We use the compatibility module in Section~\ref{sec:memory_guided_selection} to compute action scores from the updated memories. The updated memory states of the starting node and current node form $\hat{\mathbf h}_{(C)}^t$, which is used to score the unvisited nodes:
\begin{equation}
\begin{aligned}
\hat{\mathbf h}_{(C)}^t
&=\left[\widetilde{\mathbf m}_{t,\pi_1}\Vert\widetilde{\mathbf m}_{t,\pi_{t-1}}\right]W_{(C)}+\bm b_{(C)},\\
u_{\bm\theta}(i,\mathcal H_t)
&=\begin{cases}
\zeta\tanh\!\left(\hat{\mathbf h}_{(C)}^t\widetilde{\mathbf m}_{t,i}^{\mathbf T}/\sqrt d\right), & i\in\mathcal N_t,\\
-\infty, & i\in\mathcal A_t\setminus\mathcal N_t,
\end{cases}
\end{aligned}
\label{eq:pdtsp_action_scores}
\end{equation}
where $W_{(C)}\in\mathbb R^{2d\times d}$ and $\bm b_{(C)}\in\mathbb R^{1\times d}$ are learnable parameters, and $\zeta$ is the clipping parameter. The action probabilities follow Equation~\ref{eq:masked_policy}. Thus, a delivery node can contribute to the current representations before its paired pickup node is visited, while its action score is masked to $-\infty$. Selecting a pickup node makes its unvisited paired delivery node a feasible action at the next step. Construction continues until all pickup and delivery nodes have been visited, and the final edge connects the last visited node to the depot.

\subsection{Knapsack Problem}
\label{append:implementation_kp}

\subsubsection{Problem Setup}

A KP instance $\mathcal G$ consists of $N$ items and a knapsack with initial capacity $c_1$, where $c_t$ denotes the remaining capacity at construction step $t$. Each item $i\in\{1,\ldots,N\}$ has weight $w_i$ and value $v_i$. Each item may be selected at most once and cannot be divided. A feasible solution $\bm y=(y_1,\ldots,y_N)\in\Omega(\mathcal G)$ is a binary selection vector, where $y_i=1$ indicates that item $i$ is selected and $y_i=0$ otherwise. The total weight of the selected items must not exceed $c_1$, and the objective is to maximize their total value.

\subsubsection{MiLoop for KP}
\label{append:miloop_kp}
MiLoop for KP follows the memory fusion, selective memory update, and action selection procedure in Section~\ref{sec:methodology}. It constructs a solution by selecting items subject to the remaining knapsack capacity. Each item is represented by its current features and a memory state that is carried forward across construction steps.

\paragraph{Current Input Embedding and Feasibility}
Let $w_i$ and $v_i$ denote the weight and value of item $i$, respectively, and let $c_t$ be the remaining capacity at construction step $t$. The candidate set $\mathcal A_t$ contains all unselected items, and the feasible action set is $\mathcal N_t=\{i\in\mathcal A_t:w_i\leq c_t\}$. For each candidate, the model input consists of its per-unit value and its weight relative to the remaining capacity. These features are mapped to a $d$-dimensional embedding through a shared linear projection:
\begin{equation}
\bm x_{t,i}=\left[\frac{v_i}{w_i},\frac{w_i}{c_t}\right],
\qquad
\mathbf h_{t,i}^{(0)}=\bm x_{t,i}W^{(0)}+\bm b^{(0)},
\qquad i\in\mathcal A_t,
\label{eq:kp_input_embedding}
\end{equation}
where $W^{(0)}$ and $\bm b^{(0)}$ are learnable parameters. Recomputing these features at each step allows the current embeddings to reflect the remaining capacity. The resulting embeddings form $H_t^{(0)}$ and provide the current input for memory fusion.

\paragraph{Stepwise Memory Fusion and Selective Update}
We maintain a memory matrix $M_t^{(\mathcal G)}\in\mathbb R^{N\times d}$ for the $N$ items, where $\mathbf m_{t,i}$ stores the historical representation of item $i$. On the first call, the current embeddings serve directly as the decoder input and the initial memories. At subsequent steps, the memories of the unselected items are retrieved and fused with their current embeddings:
\begin{equation}
\widetilde{\mathbf h}_{t,i}^{(0)}
=\left[\mathbf m_{t,i}\Vert\mathbf h_{t,i}^{(0)}\right]W_{\mathrm{fusion}},
\qquad i\in\mathcal A_t,
\label{eq:kp_input_fusion}
\end{equation}
where $W_{\mathrm{fusion}}$ is a learnable fusion matrix shared across items. The fused embeddings are processed by stacked attention layers to obtain the current candidate representations $\widetilde H_t^{(L)}$, as in Equation~\ref{eq:miloop_shallow_decoder}. All unselected items participate in this computation, including those that exceed the remaining capacity.

The current candidate representations and historical memories then determine adaptive gates and updated memories:
\begin{equation}
\begin{aligned}
\bm g_{t,i}
&=\sigma\!\left(\left[\mathbf m_{t,i}\Vert\widetilde{\mathbf h}_{t,i}^{(L)}\right]W_g+\bm b_g\right),\\
\widetilde{\mathbf m}_{t,i}
&=\left(1-\bm{g}_{t,i}\right)\odot\mathbf m_{t,i}
+\bm{g}_{t,i}\odot\widetilde{\mathbf h}_{t,i}^{(L)}.
\end{aligned}
\label{eq:kp_memory_update}
\end{equation}
Here, $W_g$ and $\bm b_g$ are learnable gate parameters shared across items and construction steps. The updated memories provide the input for current action scoring and are carried forward by item identity for reuse in subsequent steps.

\paragraph{Memory-Guided Action Selection}
A shared linear projection maps each updated memory to a scalar action score, with infeasible items masked out:
\begin{equation}
 u_{\bm\theta}(i,\mathcal H_t)=
 \begin{cases}
  \widetilde{\mathbf m}_{t,i}W_{\mathrm{out}}+b_{\mathrm{out}}, & i\in\mathcal N_t,\\
  -\infty, & i\in\mathcal A_t\setminus\mathcal N_t,
 \end{cases}
\label{eq:kp_action_scores}
\end{equation}
where $W_{\mathrm{out}}$ and $b_{\mathrm{out}}$ are learnable parameters of the output module. A softmax yields the action probabilities, following Equation~\ref{eq:masked_policy}. Selecting item $i$ adds its value to the solution, reduces $c_t$ by $w_i$, and removes it from $\mathcal A_t$. The model then rebuilds the features of the remaining items and fuses them with their updated memories at the next step. Construction terminates when no unselected item fits the remaining capacity.

\subsection{Adaptation Attention Free Module}
\label{append:aafm}

\subsubsection{Overview}
Following \citet{zhou2026icam}, we implement $\operatorname{Attention}(\cdot)$ using an adaptation attention free module (AAFM) to enhance structural perception for COPs. It replaces dot-product attention with feature-wise aggregation guided by an instance-dependent pairwise bias. We use AAFM as the attention operator in Equation~\ref{eq:attention_layer}. Given the input $\widetilde H_t^{(\ell-1)}$, its query, key, and value projections are
\begin{equation}
Q=\widetilde H_t^{(\ell-1)}W_Q^{(\ell)},\qquad
K=\widetilde H_t^{(\ell-1)}W_K^{(\ell)},\qquad
V=\widetilde H_t^{(\ell-1)}W_V^{(\ell)},
\label{eq:aafm_projections}
\end{equation}
where $W_Q^{(\ell)},W_K^{(\ell)},W_V^{(\ell)}\in\mathbb R^{d\times d}$ are learnable matrices. Omitting step and layer indices, AAFM computes
\begin{equation}
\operatorname{AAFM}(Q,K,V;B)
=\sigma(Q)\odot
\frac{\exp(B)\bigl(\exp(K)\odot V\bigr)}{\exp(B)\exp(K)}.
\label{eq:aafm_aggregation}
\end{equation}
Here, $\sigma$ is the sigmoid function, $\odot$ denotes element-wise multiplication, and $B$ is the pairwise adaptation bias. Exponentiation and division are element-wise; products involving $\exp(B)$ are matrix multiplications. Thus, keys and biases determine normalized aggregation weights, while queries gate the output without computing $QK^{\mathbf T}$.

\subsubsection{AAFM for TSP}
For TSP, AAFM processes the starting node, current node, and unvisited nodes in $\mathcal A_t$. For unmasked input rows $i,j$, the scale--distance bias is
\begin{equation}
B_{t,ij}^{(\ell)}
=-\alpha_\ell\gamma_t\log_2|\mathcal A_t|\,
\left\|\widetilde{\bm x}_{t,i}-\widetilde{\bm x}_{t,j}\right\|_2,
\label{eq:aafm_distance_bias}
\end{equation}
where $\widetilde{\bm x}_{t,i}$ denotes normalized coordinates, $\alpha_\ell$ is learnable and initialized to $1$, and $\gamma_t=\max\{1,\log_{100}|\mathcal A_t|\}$. For positive $\alpha_\ell$, this bias favors nearby nodes while adapting to the remaining problem size. Across the four backbones, self-interactions are masked by setting $B_{t,ii}^{(\ell)}=-\infty$.

\subsubsection{AAFM for CVRP}
CVRP uses Equation~\ref{eq:aafm_distance_bias} with $\gamma_t=1$, where $\mathcal A_t$ contains unvisited customers and an available depot-return action. Demands enter the node embeddings. AAFM also refines the capacity-conditioned query using candidate memories, as in Equation~\ref{eq:cvrp_light_decoder}. Padded candidates are excluded from aggregation; capacity feasibility is enforced by the final action mask.

\subsubsection{AAFM for PDTSP}
PDTSP uses the same distance bias, with $\gamma_t=1$ during training and $\gamma_t=\max\{1,\log_{100}|\mathcal A_t|\}$ during inference. Pickup--delivery pairing enters the input embeddings and memory fusion. All unvisited nodes participate in aggregation, including deliveries whose pickups remain unvisited; pickup-before-delivery precedence is enforced through $\mathcal N_t$.

\subsubsection{AAFM for KP}
For KP, AAFM aggregates unselected items using their values $v_i$ and weights $w_i$. For distinct $i,j\in\mathcal A_t$, the bias is
\begin{equation}
\begin{aligned}
\rho_{t,ij}&=\frac{v_i+v_j}{w_i+w_j},\qquad
\rho_{t,i}^{\max}=\max_{k\in\mathcal A_t\setminus\{i\}}\rho_{t,ik},\\
B_{t,ij}^{(\ell)}&=-\alpha_\ell\log_2|\mathcal A_t|
\left(1-\frac{\rho_{t,ij}}{\rho_{t,i}^{\max}}\right).
\end{aligned}
\label{eq:aafm_kp_bias}
\end{equation}
Numerical stabilizers are omitted for clarity. For positive $\alpha_\ell$, the bias favors pairs with higher combined value per unit weight. Self-interactions and padded items are masked, while capacity feasibility is enforced during action selection.


\section{Experimental Setup}
\subsection{Problem and Hyperparameter Settings}
\label{append:model_setting}
Detailed information on the problem and hyperparameter settings can be found in \cref{tab:dataset_summary} and \cref{table:hyperparameters}, respectively. TSP, CVRP, and PDTSP models are trained on uniformly distributed instances. The training size for KP follows \citet{kwon2020pomo}. For each problem, the same trained model is used in all evaluations.

\begin{table}[h!]
\centering
\caption{Summary of the test datasets and evaluation settings.}
\label{tab:dataset_summary}

\begin{tabularx}{\linewidth}{
    l c r >{\raggedright\arraybackslash}X
}
    \toprule
    \textbf{Problem / dataset}
    & \textbf{Size $N$}
    & \textbf{\# instances}
    & \textbf{Capacity / details} \\
    \midrule

    \multicolumn{4}{l}{
        \textbf{Synthetic test sets}
    } \\

    \multirow{3}{*}{TSP}
    & $100$ & $10{,}000$ & \multirow{3}{*}{---} \\
    & $1\mathrm{K}$ & $128$ & \\
    & $5\mathrm{K},\,10\mathrm{K}$ & $16$ & \\

    \multirow{3}{*}{CVRP}
    & $100$ & $10{,}000$ & $C=50$ \\
    & $1\mathrm{K}$ & $128$ & $C=50$ \\
    & $5\mathrm{K},\,10\mathrm{K}$ & $20$ & $C=50$ \\

    \multirow{2}{*}{PDTSP}
    & $100$ & $1{,}000$ & \multirow{2}{*}{---} \\
    & $200,\,500,\,1\mathrm{K}$ & $128$ & \\

    KP
    & $200,\,500,\,1\mathrm{K}$
    & $128$
    & $C \in \{10,\,25,\,50,\,100\}$ \\

    \midrule
    \multicolumn{4}{l}{
        \textbf{Benchmark test sets}
    } \\

    \multirow{4}{*}{TSPLib~\citep{reinelt1991tsplib}}
    & $(0,\,1\mathrm{K}]$
    & $49$
    & \multirow{4}{*}{
        \shortstack[l]{
            Symmetric distance\\
            \texttt{EUC\_2D}\\
            The maximum scale is 85,900.
        }
      } \\
    & $(1\mathrm{K},\,5\mathrm{K}]$
    & $22$ & \\
    & $(5\mathrm{K},\,100\mathrm{K})$
    & $10$ & \\
    \cmidrule(lr){2-3}
    & \textbf{Total}
    & \textbf{81} & \\

    \multirow{4}{*}{CVRPLib}
    & $(0,\,1\mathrm{K}]$
    & $100$
    & Set-X~\citep{uchoa2017cvrplib_setx} \\
    & $(1\mathrm{K},\,7\mathrm{K}]$
    & $4$
    & Set-AGS~\citep{arnold2019cvrplib_xxl} \\
    & $(7\mathrm{K},\,30\mathrm{K}]$
    & $6$
    & Set-AGS~\citep{arnold2019cvrplib_xxl} \\
    \cmidrule(lr){2-4}
    & \textbf{Total}
    & \textbf{110}
    & Set-X + Set-AGS \\

    DIMACS Challenge
    & $[10\mathrm{K},\,10\mathrm{M}]$
    & $11$
    & The 8th DIMACS challenge E series TSP instances \\

    \bottomrule
\end{tabularx}
\par
\begin{minipage}{\linewidth}
    \textit{Notes.}
    $N$ denotes the instance size: nodes for routing problems
    and items for KP.
    $\mathrm{K}=10^3$ and $\mathrm{M}=10^6$.
    KP is evaluated at all four listed capacities for each size.
    Benchmark totals refer to the instances selected for this study.
\end{minipage}
\end{table}

\begin{table}[h!]
\centering
\caption{Model and training settings of MiLoop.}
\label{table:hyperparameters}
\resizebox{0.8\linewidth}{!}{
\begin{tabular}{lcccc}
\toprule[0.5mm]
\textbf{Setting}
& TSP
& CVRP
& PDTSP
& KP \\
\midrule
\multicolumn{5}{l}{\textit{Model settings}} \\
Embedding dimension $d$
& \multicolumn{4}{c}{$128$} \\
Feed-forward hidden dimension
& \multicolumn{4}{c}{$512$} \\
Number of attention layers $L$
& \multicolumn{4}{c}{$3$} \\
Attention mechanism
& \multicolumn{4}{c}{AAFM~\citep{zhou2026icam}} \\
Logit clipping parameter $\xi$
& \multicolumn{4}{c}{$10$} \\
\midrule
\multicolumn{5}{l}{\textit{Optimization settings}} \\
Optimizer
& \multicolumn{4}{c}{AdamW} \\
Initial learning rate
& \multicolumn{4}{c}{$10^{-4}$} \\
Learning rate decay factor
& \multicolumn{4}{c}{$0.1$} \\
Learning rate decay epoch
& $101$& $101$& $101$
& $15$ \\
Preference scaling factor $\beta$
& \multicolumn{4}{c}{$0.1$} \\
\midrule
\multicolumn{5}{l}{\textit{Training settings}} \\
Training instance size
& $100$
& $100$
& $100$
& $200$ \\
Batch size
& $120$
& $100$
& $120$
& $40$ \\
Capacity
& $-$
& $50$
& $-$
& $25$ \\
Gradient clipping
& \multicolumn{4}{c}{$1.0$} \\
Batches per epoch
& \multicolumn{4}{c}{$2{,}500$} \\
Total epochs
& $110$& $110$& $110$
& $20$ \\
\bottomrule[0.5mm]
\end{tabular}
}

\end{table}

\subsection{Pseudocode for Training}
\label{append:dpo}

As shown in Algorithm~\ref{alg:miloop_dpo_training}, we detail the reference-free DPO procedure of MiLoop, illustrated on the TSP. Each batch samples two tours per instance under the current policy and performs one optimizer update. We use the notation of Section~\ref{sec:miloop_training}: $B$ is the effective batch size after any data augmentation, $T=N$ is the tour length, and $\ell_b^{(k)}$ accumulates the sequence log-probability for sample $k\in\{1,2\}$ of instance $\mathcal G_b$. Batch and sample indices are suppressed inside the construction loop for readability.

The two trajectories of each instance have separate environments and memories but share $\bm\theta$. Sampling and preference ordering are treated as fixed when differentiating the loss, while the computational graph through the action log-probabilities and memory updates is retained throughout the rollout. Equal-cost pairs follow the implementation's tie rule: the second sampled tour is assigned to $\bm\pi_b^+$. The loss is evaluated with a numerically stable log-sigmoid, and neither a reference-policy likelihood nor a reward baseline enters Equation~\ref{eq:miloop_dpo_loss}.

The Algorithm~\ref{alg:miloop_dpo_training} describes the default configuration with one update per batch. When multiple updates per collected pair are configured, the implementation stores the sampled actions and replays them with reset environments and memories to recompute their log-probabilities under the current parameters before each update.

\begin{algorithm}[H]
\caption{Reference-free DPO training of MiLoop}
\label{alg:miloop_dpo_training}
\small
\begin{algorithmic}[1]
\STATE {\bfseries Input:} Number of epochs $E$, steps per epoch $S$, decision steps $T$, batch size $B$, preference scale $\beta>0$.
\STATE {\bfseries Output:} The trained model with parameters $\bm\theta=\{\bm\theta_{\mathrm{init}},\bm\theta_{\mathrm{fusion}},\bm\theta_{\mathrm{attn}},\bm\theta_{\mathrm{update}},\bm\theta_{\mathrm{logit}}\}$.
\STATE Initialize AdamW and $\bm\theta$.
\FOR{$\text{epoch} = 1, \ldots, E$}
    \FOR{$\text{step} = 1, \ldots, S$}
        \STATE Set the policy to training mode with stochastic decoding; clear parameter gradients.
        \STATE Sample $\{\mathcal G_b\}_{b=1}^{B}$ and duplicate them to form a batch of $2B$ trajectories.
        \STATE For each copy, reset its environment and memory; set $\pi_1\gets0$ and $\ell\gets0$.
        \STATE Visit node $0$ and initialize the remaining-node state $s_2$.
        \FOR{$t=2,\ldots,T$}
            \FORALL{$(b,k)\in\{1,\ldots,B\}\times\{1,2\}$ in parallel}
                \STATE Construct $\mathcal H_t$, feasible nodes $\mathcal N_t$, and current input embeddings $H_t^{(0)}$.
                \IF{$t=2$}
                    \STATE Set $\widetilde H_t^{(0)}\gets H_t^{(0)}$; use current role embeddings as initial memories.
                \ELSE
                    \STATE Retrieve memories from $M_t^{(\mathcal G)}$ and fuse them with $H_t^{(0)}$ using Equation~\ref{eq:miloop_input_fusion}.
                \ENDIF
                \STATE Compute $\widetilde H_t^{(L)}$ with the attention layers using Equation~\ref{eq:miloop_shallow_decoder}.
                \STATE Compute gates and updated memories $\widetilde M_t^{(\mathcal G)}$ using Equations~\ref{eq:miloop_memory_gate}--\ref{eq:miloop_memory_update}.
                \STATE Compute $p_{\bm\theta}(\cdot\mid\mathcal H_t)$ from the updated memories; sample $\pi_t\sim p_{\bm\theta}(\cdot\mid\mathcal H_t)$.
                \STATE $\ell\gets\ell+\log\max\{p_{\bm\theta}(\pi_t\mid\mathcal H_t)\}$.
                \STATE Write updated memories back by node identity and carry them to $M_{t+1}^{(\mathcal G)}$.
                \STATE Append $\pi_t$ to the tour and update $s_{t+1}$ by marking the selected node visited.
            \ENDFOR
        \ENDFOR
        \STATE Obtain $\bm\pi_b^{(k)}$, $\ell_b^{(k)}$, and $f(\bm y_b^{(k)}\mid\mathcal G_b)$, including the closing edge to node $0$.
        \FOR{$b=1,\ldots,B$}
            \IF{$f(\bm y_b^{(1)}\mid\mathcal G_b)<f(\bm y_b^{(2)}\mid\mathcal G_b)$}
                \STATE $(\bm\pi_b^+,\ell_b^+)\gets(\bm\pi_b^{(1)},\ell_b^{(1)})$;\quad $(\bm\pi_b^-,\ell_b^-)\gets(\bm\pi_b^{(2)},\ell_b^{(2)})$.
            \ELSE
                \STATE $(\bm\pi_b^+,\ell_b^+)\gets(\bm\pi_b^{(2)},\ell_b^{(2)})$;\quad $(\bm\pi_b^-,\ell_b^-)\gets(\bm\pi_b^{(1)},\ell_b^{(1)})$.
            \ENDIF
        \ENDFOR
        \STATE $\displaystyle \mathcal L(\bm\theta)\gets-\frac{1}{B}\sum_{b=1}^{B}\log\sigma\!\left(\beta[\ell_b^+-\ell_b^-]\right)$.
        \STATE $\bm{\theta} \gets \text{AdamW}(\bm{\theta}, \nabla\mathcal{L(\bm\theta)})$
        
    \ENDFOR
\ENDFOR
\end{algorithmic}
\end{algorithm}

\clearpage

\section{Effectiveness of Memory}
\label{append:memory_effectiveness}

\paragraph{Experimental setup}
To examine stepwise memory propagation, we compare MiLoop with a traditional dynamic embedding baseline that recomputes dynamic node embeddings at each construction step. During greedy decoding on TSP100, we record the embeddings of all unvisited nodes immediately before action scoring at each step and treat them collectively as the model's representation of the current state. For MiLoop, these embeddings are extracted after information fusion and selective updates, thereby incorporating historical information from preceding construction steps.

\paragraph{Similarity measurements}
We use cosine similarity to quantify how node representations evolve during construction and examine the role of memory in this evolution.
Consecutive subproblems share most of their unvisited nodes, so high similarity can partly reflect a common component across node embeddings.
To distinguish this shared component from the continuity of individual node representations, we center the embeddings before comparing adjacent steps.
Following the notation in the main text, let $z_{t,i}\in\mathbb{R}^{d}$ denote the representation of an unvisited node $i\in\mathcal A_t$, taken from $H_t^{(L)}$ for the baseline or $\widetilde M_t^{(\mathcal G)}$ for MiLoop.
For an instance of size $N$, the $N-1$ decisions after selecting the starting node $\pi_1$ correspond to construction steps $t=2,\ldots,N$.
We match nodes by identity over $I_t=\mathcal A_{t-1}\cap\mathcal A_t=\mathcal A_t$ and subtract the mean embedding at each step, computed over this same set:
\begin{equation}
\widetilde z_{s,i}^{(t)}=z_{s,i}-\mu_s^{(t)},
\qquad
\mu_s^{(t)}=\frac{1}{|I_t|}\sum_{j\in I_t}z_{s,j},
\qquad s\in\{t-1,t\}.
\label{eq:memory_heatmap_centering}
\end{equation}
This removes the mean component shared by the remaining nodes and expresses each node's representation relative to its peers.

We compare each unvisited node's representation with two references: its representation at the preceding step, yielding $C_{t,i}$, and its representation at the first recorded step, yielding $A_{t,i}$:
\begin{equation}
\begin{aligned}
C_{t,i}&=\operatorname{cos}\!\left(\widetilde z_{t-1,i}^{(t)},\;\widetilde z_{t,i}^{(t)}\right),\\
A_{t,i}&=\operatorname{cos}(z_{t,i},z_{2,i}).
\end{aligned}
\label{eq:memory_heatmap_measures}
\end{equation}
Here, $\operatorname{cos}(a,b)=a^\top b/(\lVert a\rVert_2\lVert b\rVert_2)$.
The adjacent-step similarity $C_{t,i}$ captures continuity in each node's relative representation, while the first-step similarity $A_{t,i}$ tracks directional change from its initial representation over the rollout.

\begin{figure}[H]
    \centering
    \captionsetup[subfigure]{font=small,skip=3pt,justification=centering}
    \begin{subfigure}[t]{0.3485\linewidth}
        \centering
        \includegraphics[width=\linewidth]{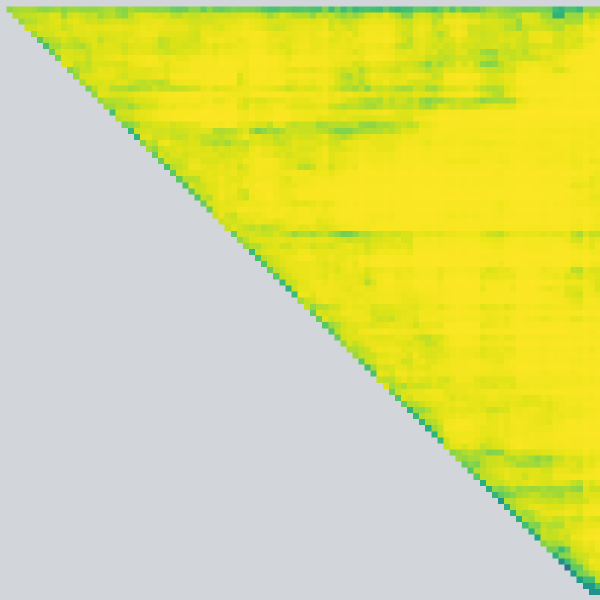}
        \caption{MiLoop: adjacent-step cosine}
        \label{fig:miloop_k1}
    \end{subfigure}\hspace{0.0255\linewidth}%
    \begin{subfigure}[t]{0.3485\linewidth}
        \centering
        \includegraphics[width=\linewidth]{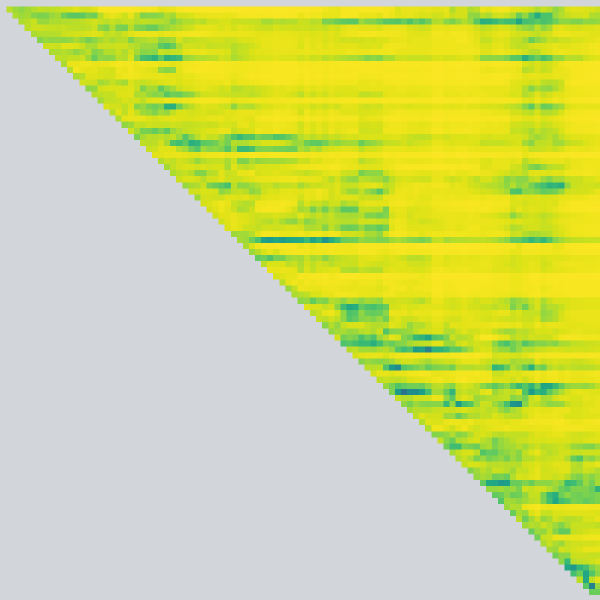}
        \caption{Dynamic baseline: adjacent-step cosine}
        \label{fig:decoder_k1}
    \end{subfigure}

    \par\vspace{6.8pt}
    \begin{subfigure}[t]{0.3485\linewidth}
        \centering
        \includegraphics[width=\linewidth]{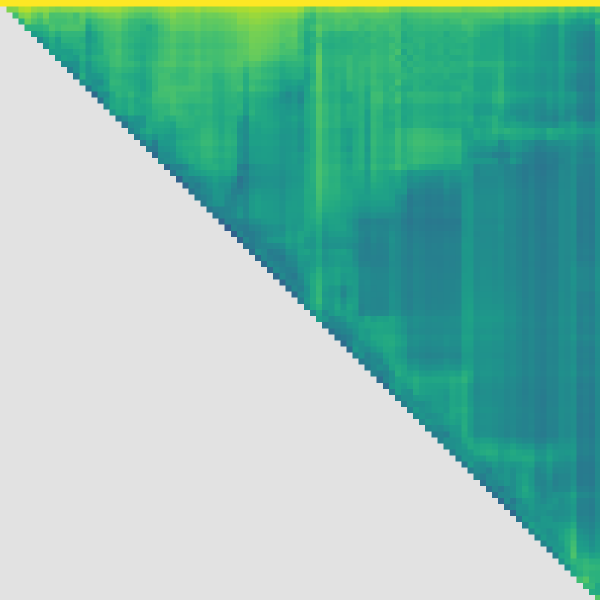}
        \caption{MiLoop: first-step cosine}
        \label{fig:miloop_k100}
    \end{subfigure}\hspace{0.0255\linewidth}%
    \begin{subfigure}[t]{0.3485\linewidth}
        \centering
        \includegraphics[width=\linewidth]{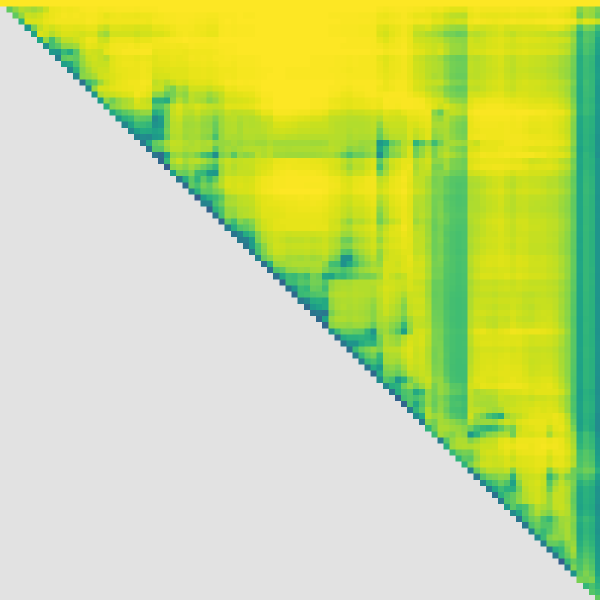}
        \caption{Dynamic baseline: first-step cosine}
        \label{fig:decoder_k100}
    \end{subfigure}
    \caption{Node-representation similarity during greedy TSP100 construction. The left column shows MiLoop and the right column shows the traditional dynamic-embedding baseline. Figures (a,b) report adjacent-step cosine $C_{t,i}$; Figures (c,d) report cosine to the first recorded representation $A_{t,i}$. Within each heatmap, rows correspond to autoregressive construction steps, and columns correspond to the nodes that remain unvisited at that step. Colors closer to yellow indicate higher similarity, while colors closer to green indicate lower similarity. As construction proceeds, visited nodes drop out of subsequent comparisons, producing the shrinking triangular region; the unvisited set is exhausted when the tour is complete. Gray entries indicate visited nodes or undefined similarities.}
    \label{fig:heatmap_comparison}
\end{figure}

Table~\ref{tab:memory_similarity_means} shows the average similarities of MiLoop and 3-layer transformer baseline on two different similarities, and the detailed reults are demonstrated in Figure~\ref{fig:heatmap_comparison}.

\begin{table}[H]
    \centering
    \small
    \caption{Mean cosine similarities of adjacent steps and first-recorded step.}
    \label{tab:memory_similarity_means}
    \begin{tabular}{lrr}
        \toprule
        Similarity measure & Traditional dynamic embedding & MiLoop \\
        \midrule
        Adjacent steps ($C_{t,i}$) & 0.8480 & 0.9159 \\
        First recorded step ($A_{t,i}$) & 0.7655 & 0.4497 \\
        \bottomrule
    \end{tabular}
\end{table}

\paragraph{Results}

For adjacent steps, MiLoop achieves a mean similarity of $0.9159$, compared with $0.8480$ for the traditional dynamic embedding, a difference of $0.0679$.
MiLoop's higher adjacent-step similarity, together with the more uniform pattern in Figure~\ref{fig:heatmap_comparison}\subref{fig:miloop_k1}, supports the interpretation that memory propagation preserves continuity across successive decisions, allowing its state representations to evolve smoothly.
In contrast, the baseline recomputes node embeddings without memory propagation and exhibits less consistent similarity, with pronounced bands of reduced similarity in Figure~\ref{fig:heatmap_comparison}\subref{fig:decoder_k1} indicating more abrupt transitions in its state representations.

When the first recorded step serves as the reference, MiLoop's mean similarity is $0.4497$, compared with $0.7655$ for the baseline, a difference of $0.3158$.
Figures~\ref{fig:heatmap_comparison}\subref{fig:decoder_k100} and~\ref{fig:heatmap_comparison}\subref{fig:miloop_k100} show how this difference develops: the baseline remains broadly aligned with its initial embeddings, whereas MiLoop departs from them early and maintains lower similarity through much of construction.
The lower mean therefore reflects a sustained departure from the initial representational directions across the remaining nodes.

\begin{figure}[H]
    \centering
    \includegraphics[width=0.90\linewidth]{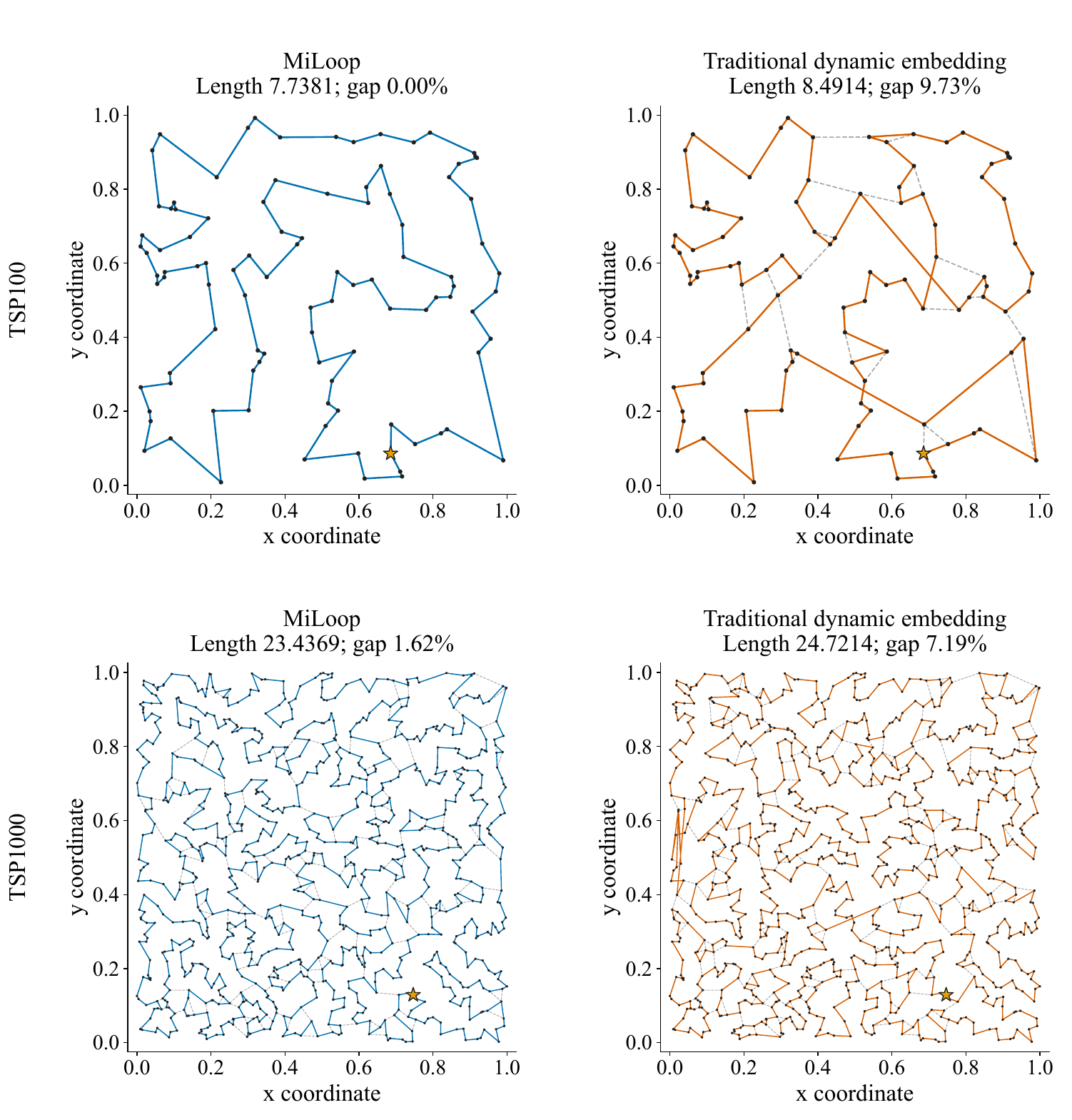}
    \caption{Tour comparisons between MiLoop (left) and the traditional dynamic embedding (right) on TSP100 (top) and TSP1000 (bottom). Solid blue and orange lines depict the tours produced by MiLoop and the baseline, respectively. Black dots denote nodes, and stars mark the starting nodes.}
    \label{fig:memory_solution_visualization}
\end{figure}

\clearpage
\paragraph{Solution visualization}
To complement the representation analysis, Figure~\ref{fig:memory_solution_visualization} compares the tours produced by MiLoop and the traditional dynamic embedding on TSP100 and TSP1000.
On the TSP100 example, MiLoop achieves a gap of $0.00\%$, compared with $9.73\%$ for the baseline.
On TSP1000, MiLoop also obtains a shorter tour, with a gap of $1.62\%$ versus $7.19\%$.
These examples provide a direct comparison of solution quality alongside the preceding analysis of representation dynamics.

Across these comparisons, MiLoop exhibits stronger continuity between successive decisions alongside a larger accumulated departure from its initial representations.
This combination is consistent with selective memory propagation, in which information carried forward supports further updates as the construction state evolves.
The route visualizations complement these observations with lower gaps at both scales.
\clearpage

\section{Experiment Details}
\subsection{Results on Synthetic PDTSP and KP}

\label{append:results_uniform_pdtsp_kp}
\begin{table}[htbp]
  \centering
  \caption{Comparison on PDTSP instances with $N\leq1{,}000$. }
    \resizebox{\textwidth}{!}{
    \begin{threeparttable}
    \begin{tabular}{l|ccc|ccc|ccc|ccc}
    \toprule[0.5mm]
    \multirow{2}[2]{*}{Method} & \multicolumn{3}{c|}{PDTSP100} & \multicolumn{3}{c|}{PDTSP200} & \multicolumn{3}{c|}{PDTSP500} & \multicolumn{3}{c}{PDTSP1000} \\
          & Obj.  & Gap   & Time  & Obj.  & Gap   & Time  & Obj.  & Gap   & Time  & Obj.  & Gap   & Time \\
    \midrule
    LKH3  & 9.43 & 0.00\% & 5.88s & 13.47 & 0.00\% & 28.26s & 21.54  & 0.00\% & 2.06m & 31.24 & 0.00\% & 5.86m \\
    Heter-AM$\dagger$ & 10.35 & 9.76\% & 0.002s &    $-$   &   $-$    &   $-$    &    $-$   &   $-$    &   $-$    &   $-$    &   $-$    &  \\
    URS aug $\times$ 8  & 9.90  & 4.98\% & 0.004s & 14.47 & 7.42\% & 0.05s & 24.27 & 12.67\% & 0.38s & 35.92 & 14.98\% & 2.87s \\
    \midrule
    MiLoop greedy& 9.86  & 4.56\% & 0.005s & 13.75 & 2.09\% & 0.02s & 21.92 & 1.74\% & 0.16s & 31.98 & 2.40\% & 1.09s \\
    MiLoop ($M=100$) & \greybg{9.68} & \greybg{2.67\%} & 0.292s & \greybg{13.51} & \greybg{0.29\%} & 1.37s & \greybg{21.48} & \greybg{-0.27\%} & 15.17s & \greybg{31.14} & \greybg{-0.35\%} & 1.75m \\
    \bottomrule[0.5mm]
    \end{tabular}%
    \begin{tablenotes}
\item[$\dagger$] The Heter-AM results for $N=100$ are taken from ~\citep{ma2022efficient} due to the unavailability of a pretrained model of its official implementation.
\end{tablenotes}
\end{threeparttable}
}
  \label{tab:uniform_pdtsp}%
\end{table}%

As shown in~\cref{tab:uniform_pdtsp}, even with greedy decoding ($M=1$), MiLoop outperforms the evaluated neural baselines across all tested scales from $100$ to $1\mathrm{K}$ nodes, with its advantage over URS becoming more pronounced as the instance size increases.
On PDTSP500 and PDTSP1K, MiLoop achieves gaps of $1.74\%$ and $2.40\%$, respectively, compared with $12.67\%$ and $14.98\%$ for URS with $\times 8$ augmentation.
Multi-start decoding ($M=100$) further improves solution quality at all scales, achieving gaps of $-0.27\%$ and $-0.35\%$ relative to LKH3 on PDTSP500 and PDTSP1K, respectively.
Notably, MiLoop is trained only on $100$-node instances and uses the same model at all evaluation sizes.
These results demonstrate strong cross-scale generalization on routing problems with pickup-before-delivery constraints.

\begin{table}[htbp]
\centering
\caption{Comparison on KP with scale $\leq 1,000$ across different capacity settings.}
\resizebox{\textwidth}{!}{\begin{tabular}{c|c|c|crc|ccc|ccc}
\toprule[0.5mm]
\multirow{2}[2]{*}{} & \multirow{2}[2]{*}{Capacity} & OR-Tools & \multicolumn{3}{c|}{POMO (all Trajec.)} & \multicolumn{3}{c|}{BQ (greedy)} & \multicolumn{3}{c}{MiLoop (greedy)} \\
& & Value & Value & \multicolumn{1}{c}{Gap} & Time & Value & Gap & Time & Value & Gap & Time \\
\midrule
\multirow{4}[1]{*}{KP200} & C=10 & 36.0240 & 34.9182 & 3.070\% & 1s & 35.9136 & 0.309\% & 4s & \greybg{35.9559} & \greybg{0.189\%} & 2s \\
& C=25 & 57.1729 & \greybg{57.1646} & \greybg{0.014\%} & 1s & 57.1184 & 0.096\% & 4s & 57.1271 & 0.080\% & 3s \\
& C=50 & 80.7133 & 79.8295 & 1.095\% & 1s & 80.1901 & 0.650\% & 5s & \greybg{80.6814} & \greybg{0.040\%} & 3s \\
& C=100 & 99.4987 & 99.2055 & 0.295\% & 1s & 99.4338 & 0.065\% & 5s & \greybg{99.4945} & \greybg{0.004\%} & 4s \\
\midrule
\multirow{4}[1]{*}{KP500} & C=10 & 57.4130 & 54.3516 & 5.332\% & 3s & 56.8149 & 1.045\% & 21s & \greybg{57.2083} & \greybg{0.357\%} & 3s \\
& C=25 & 90.6972 & 86.3773 & 4.763\% & 3s & 90.3930 & 0.336\% & 29s & \greybg{90.6319} & \greybg{0.072\%} & 4s \\
& C=50 & 128.369 & 128.3156 & 0.042\% & 5s & 128.2786 & 0.070\% & 37s & \greybg{128.325} & \greybg{0.034\%} & 6s \\
& C=100 & 181.7669 & 181.2869 & 0.264\% & 7s & 181.0279 & 0.408\% & 44s & \greybg{181.7202} & \greybg{0.026\%} & 10s \\
\midrule
\multirow{4}[2]{*}{KP1000} & C=10 & 81.5516 & 57.8980 & 29.005\% & 8s & 79.9902 & 1.923\% & 1.9m & \greybg{80.8574} & \greybg{0.851\%} & 10s \\
& C=25 & 129.1877 & 124.0267 & 3.995\% & 15s & 128.4923 & 0.540\% & 2.8m & \greybg{129.0891} & \greybg{0.076\%} & 12s \\
& C=50 & 182.7367 & 172.2880 & 5.718\% & 20s & 181.9492 & 0.430\% & 3.6m & \greybg{182.6674} & \greybg{0.038\%} & 19s \\
& C=100 & 258.252 & 254.5366 & 1.439\% & 28s & 258.0553 & 0.076\% & 4.4m & \greybg{258.2178} & \greybg{0.013\%} & 25s \\
\bottomrule[0.5mm]
\end{tabular}}%
\label{tab:uniform_kp}%
\end{table}%

Table~\ref{tab:uniform_kp} reports results across different problem sizes and knapsack capacities.
Even with greedy decoding, MiLoop achieves the lowest gaps among neural solvers in $11$ of the $12$ settings, including all settings with $500$ and $1\mathrm{K}$ items.
On KP1K, MiLoop maintains gaps below $1\%$ across all tested capacities.
At $C=10$, it achieves a gap of $0.851\%$, compared with $1.923\%$ for BQ and $29.005\%$ for POMO; at $C=100$, its gap decreases to $0.013\%$.
MiLoop also requires only $10$--$25$\,s on KP1K, compared with $1.9$--$4.4$\,min for BQ.
The only setting where MiLoop does not achieve the lowest neural-solver gap is KP200 with $C=25$, where POMO achieves $0.014\%$ compared with MiLoop's $0.080\%$.
Overall, MiLoop maintains high solution quality on larger instances across varied capacities despite being trained only on $200$-item instances, extending its strong cross-scale generalization beyond routing problems.
\clearpage

\subsection{Results on TSPLIB and CVRPLIB }
\label{append:tsplib_cvrplib_results}

We provide detailed results on individual TSPLib and CVRPLib instances to complement the aggregate comparison in Tables~\ref{tab:benchmark}.
The evaluation covers symmetric \texttt{EUC\_2D} instances from TSPLib~\citep{reinelt1991tsplib} and from CVRPLib (Set-X~\citep{uchoa2017cvrplib_setx} and Set-AGS~\citep{arnold2019cvrplib_xxl}), spanning a broad range of problem sizes and spatial distributions.
Tables~\ref{tab:tsplib_dynamic} and~\ref{tab:cvrplib_dynamic} report the solution gap for each instance, comparing MiLoop under greedy and multi-start decoding ($M=100$) with constructive neural solvers using dynamic embeddings.
For each problem, MiLoop uses the same model trained on uniformly distributed $100$-node instances, allowing these comparisons to assess generalization across both problem scales and data distributions, as well as the benefit of multi-start decoding.
\begingroup
\footnotesize
\setlength{\tabcolsep}{1pt}
\renewcommand{\arraystretch}{1.04}
\setlength{\LTcapwidth}{\textwidth}
\begin{longtable}{@{}l|r|r|r|r|r|r|r|r|r|r@{}}
\caption{Detail results on TSPLIB: Constructive NCO with Dynamic Embeddings.}\label{tab:tsplib_dynamic}\\
\toprule[0.8pt]
\multicolumn{1}{c|}{Instance} & \multicolumn{1}{c|}{Scale} & \multicolumn{1}{c|}{\shortstack{BQ\\greedy}} & \multicolumn{1}{c|}{\shortstack{LEHD\\greedy}} & \multicolumn{1}{c|}{\shortstack{SIGD\\greedy}} & \multicolumn{1}{c|}{\shortstack{L2C-Insert\\greedy}} & \multicolumn{1}{c|}{\shortstack{INViT-3V\\greedy}} & \multicolumn{1}{c|}{\shortstack{L2R\\greedy}} & \multicolumn{1}{c|}{\shortstack{Rec-NCO\\greedy}} & \multicolumn{1}{c|}{\shortstack{MiLoop\\greedy}} & \multicolumn{1}{c}{\shortstack{MiLoop\\($M=100$)}} \\
\midrule[0.4pt]
\endfirsthead
\caption[]{Detail results on TSPLIB: Constructive NCO with Dynamic Embeddings. (continued)}\\
\toprule[0.8pt]
\multicolumn{1}{c|}{Instance} & \multicolumn{1}{c|}{Scale} & \multicolumn{1}{c|}{\shortstack{BQ\\greedy}} & \multicolumn{1}{c|}{\shortstack{LEHD\\greedy}} & \multicolumn{1}{c|}{\shortstack{SIGD\\greedy}} & \multicolumn{1}{c|}{\shortstack{L2C-Insert\\greedy}} & \multicolumn{1}{c|}{\shortstack{INViT-3V\\greedy}} & \multicolumn{1}{c|}{\shortstack{L2R\\greedy}} & \multicolumn{1}{c|}{\shortstack{Rec-NCO\\greedy}} & \multicolumn{1}{c|}{\shortstack{MiLoop\\greedy}} & \multicolumn{1}{c}{\shortstack{MiLoop\\($M=100$)}} \\
\midrule[0.4pt]
\endhead
\midrule[0.4pt]
\multicolumn{11}{r}{\textit{Continued on the next page.}}\\
\endfoot
\bottomrule[0.8pt]
\endlastfoot
eil51 & 51 & 0.94\% & 1.17\% & 0.47\% & 0.94\% & 3.99\% & 0.70\% & 1.41\% & 2.82\% & \greybg{0.00\%} \\
berlin52 & 52 & \greybg{0.00\%} & \greybg{0.00\%} & \greybg{0.00\%} & 0.01\% & 9.64\% & 0.77\% & \greybg{0.00\%} & \greybg{0.00\%} & \greybg{0.00\%} \\
st70 & 70 & \greybg{0.00\%} & \greybg{0.00\%} & 0.59\% & 0.15\% & 2.37\% & 0.30\% & \greybg{0.00\%} & 1.63\% & \greybg{0.00\%} \\
eil76 & 76 & \greybg{0.00\%} & 1.49\% & 0.56\% & 0.56\% & 5.76\% & 3.53\% & 0.93\% & 0.19\% & \greybg{0.00\%} \\
pr76 & 76 & 0.92\% & 0.22\% & 0.14\% & 1.62\% & 5.90\% & 2.76\% & 0.79\% & 1.24\% & \greybg{0.02\%} \\
rat99 & 99 & 0.58\% & 0.58\% & 0.17\% & \greybg{0.00\%} & 7.68\% & 4.54\% & 1.57\% & 2.56\% & \greybg{0.00\%} \\
kroA100 & 100 & 0.11\% & 0.11\% & 0.11\% & 0.06\% & 1.75\% & 1.25\% & 0.06\% & 0.11\% & \greybg{0.00\%} \\
kroB100 & 100 & 0.58\% & 0.26\% & 0.23\% & 0.19\% & 3.67\% & 2.16\% & \greybg{0.00\%} & 0.39\% & 0.09\% \\
kroC100 & 100 & \greybg{0.00\%} & 0.32\% & 0.01\% & 0.11\% & 8.01\% & 1.95\% & \greybg{0.00\%} & 1.13\% & 0.01\% \\
kroD100 & 100 & \greybg{0.00\%} & 0.38\% & 0.38\% & 0.07\% & 6.73\% & 2.19\% & \greybg{0.00\%} & 0.54\% & \greybg{0.00\%} \\
kroE100 & 100 & 0.31\% & 0.43\% & 0.24\% & 0.24\% & 9.88\% & 1.15\% & 0.33\% & 0.34\% & \greybg{0.19\%} \\
rd100 & 100 & \greybg{0.00\%} & \greybg{0.00\%} & 0.01\% & 0.05\% & 7.28\% & 3.74\% & \greybg{0.00\%} & 0.43\% & \greybg{0.00\%} \\
eil101 & 101 & 1.27\% & 0.64\% & 0.32\% & \greybg{0.00\%} & 3.34\% & 2.70\% & \greybg{0.00\%} & 1.43\% & \greybg{0.00\%} \\
lin105 & 105 & \greybg{0.00\%} & 0.31\% & \greybg{0.00\%} & 0.31\% & 9.14\% & 0.92\% & \greybg{0.00\%} & 0.32\% & \greybg{0.00\%} \\
pr107 & 107 & 14.01\% & 11.81\% & 14.24\% & 14.01\% & \greybg{2.14\%} & 17.04\% & 15.87\% & 19.37\% & 12.03\% \\
pr124 & 124 & 1.42\% & 1.11\% & 1.42\% & \greybg{0.00\%} & 9.08\% & 2.56\% & 0.65\% & 1.45\% & 0.08\% \\
bier127 & 127 & 4.52\% & 5.63\% & 0.19\% & 0.54\% & 16.49\% & 13.04\% & 1.33\% & 2.00\% & \greybg{0.03\%} \\
ch130 & 130 & 0.57\% & 0.57\% & 0.41\% & 1.36\% & 4.98\% & 3.99\% & 0.41\% & 1.46\% & \greybg{0.00\%} \\
pr136 & 136 & 0.01\% & 0.45\% & 0.22\% & 0.60\% & 8.43\% & 4.28\% & \greybg{0.00\%} & 2.33\% & 0.47\% \\
pr144 & 144 & 0.90\% & 3.59\% & \greybg{0.03\%} & 0.06\% & 6.89\% & 7.75\% & 0.12\% & 2.07\% & 0.05\% \\
ch150 & 150 & 0.77\% & 0.51\% & 0.93\% & 0.72\% & 4.38\% & 4.15\% & 0.61\% & 0.57\% & \greybg{0.00\%} \\
kroA150 & 150 & 0.39\% & 1.40\% & 13.76\% & 0.43\% & 4.61\% & 3.71\% & 0.46\% & 0.64\% & \greybg{0.00\%} \\
kroB150 & 150 & 0.04\% & 0.76\% & \greybg{0.02\%} & 0.27\% & 5.38\% & 0.82\% & 0.04\% & 0.57\% & \greybg{0.02\%} \\
pr152 & 152 & 4.45\% & 12.13\% & 9.40\% & \greybg{1.24\%} & 16.51\% & 26.55\% & 9.18\% & 16.77\% & 7.51\% \\
u159 & 159 & 0.75\% & 1.45\% & 1.04\% & 0.60\% & 2.00\% & 4.45\% & 0.00\% & 1.34\% & \greybg{0.00\%} \\
rat195 & 195 & 0.47\% & 1.08\% & 0.17\% & 2.76\% & 5.73\% & 3.01\% & 1.08\% & 1.38\% & \greybg{0.13\%} \\
d198 & 198 & 17.40\% & 9.07\% & 10.79\% & \greybg{2.66\%} & 16.36\% & 12.98\% & 11.43\% & 23.92\% & 7.36\% \\
kroA200 & 200 & 0.64\% & 0.63\% & 0.53\% & \greybg{0.13\%} & 10.28\% & 5.18\% & 0.39\% & 2.76\% & 0.26\% \\
kroB200 & 200 & 0.28\% & 0.14\% & 0.47\% & 1.69\% & 3.61\% & 3.05\% & \greybg{0.01\%} & 0.95\% & 0.18\% \\
ts225 & 225 & \greybg{0.00\%} & 0.28\% & 0.19\% & 36.56\% & 9.37\% & 4.66\% & 1.20\% & 3.59\% & 0.83\% \\
tsp225 & 225 & 1.17\% & 2.02\% & 1.99\% & 1.86\% & 8.96\% & 6.28\% & 2.86\% & 2.53\% & \greybg{1.05\%} \\
pr226 & 226 & 19.77\% & 0.86\% & 1.30\% & 0.17\% & 9.81\% & 17.65\% & 0.15\% & 39.31\% & \greybg{0.15\%} \\
gil262 & 262 & 0.38\% & 1.18\% & 1.14\% & 1.68\% & 6.39\% & 3.53\% & \greybg{0.08\%} & 0.80\% & 0.25\% \\
pr264 & 264 & 15.46\% & 5.17\% & 4.59\% & 3.59\% & 18.18\% & 19.38\% & \greybg{0.09\%} & 19.15\% & 6.84\% \\
a280 & 280 & 3.14\% & 2.56\% & 1.01\% & 2.79\% & 9.11\% & 6.71\% & 1.55\% & 4.54\% & \greybg{0.19\%} \\
pr299 & 299 & 2.02\% & 2.89\% & 1.57\% & 0.69\% & 9.39\% & 13.16\% & 0.31\% & 1.77\% & \greybg{0.25\%} \\
lin318 & 318 & 0.32\% & 1.38\% & 2.10\% & 1.29\% & 6.23\% & 5.46\% & \greybg{0.00\%} & 2.66\% & 0.72\% \\
rd400 & 400 & 0.48\% & 1.03\% & 0.87\% & 4.33\% & 4.38\% & 3.19\% & 0.54\% & 1.52\% & \greybg{0.17\%} \\
fl417 & 417 & 21.60\% & \greybg{4.59\%} & 6.45\% & 12.49\% & 20.82\% & 15.24\% & 6.87\% & 9.10\% & 5.27\% \\
pr439 & 439 & 2.59\% & 3.37\% & 1.97\% & 7.00\% & 12.42\% & 11.68\% & 3.30\% & 7.42\% & \greybg{0.60\%} \\
pcb442 & 442 & 2.01\% & 3.10\% & 0.59\% & 4.38\% & 4.56\% & 5.46\% & 1.17\% & 0.45\% & \greybg{0.23\%} \\
d493 & 493 & 8.29\% & 9.47\% & 6.83\% & 11.97\% & 5.19\% & 3.37\% & 6.19\% & 7.13\% & \greybg{1.75\%} \\
u574 & 574 & 1.67\% & 2.66\% & 2.12\% & 1.53\% & 19.74\% & 8.03\% & 1.64\% & 2.27\% & \greybg{0.99\%} \\
rat575 & 575 & 2.24\% & 2.70\% & 1.48\% & 3.65\% & 6.26\% & 3.04\% & 1.52\% & 1.89\% & \greybg{1.39\%} \\
p654 & 654 & \greybg{5.53\%} & 10.21\% & 28.35\% & 14.50\% & 20.41\% & 9.10\% & 60.53\% & 19.96\% & 6.71\% \\
d657 & 657 & 6.44\% & 8.07\% & 5.65\% & 3.34\% & 8.35\% & 5.92\% & 3.56\% & 3.50\% & \greybg{1.53\%} \\
u724 & 724 & 2.34\% & 3.29\% & 1.68\% & 3.77\% & 6.01\% & 3.27\% & 1.93\% & 8.52\% & \greybg{1.26\%} \\
rat783 & 783 & 3.52\% & 3.92\% & 2.23\% & 4.92\% & 5.97\% & 4.01\% & 3.46\% & 3.18\% & \greybg{1.40\%} \\
dsj1000 & 1,000 & 6.55\% & 7.89\% & 7.61\% & 8.16\% & 13.28\% & 13.79\% & 7.52\% & 8.16\% & \greybg{4.03\%} \\
pr1002 & 1,002 & 4.48\% & 4.43\% & 2.94\% & 7.31\% & 8.84\% & 5.34\% & 4.54\% & 2.01\% & \greybg{1.59\%} \\
u1060 & 1,060 & 8.95\% & 8.93\% & 7.55\% & 9.86\% & 23.75\% & 5.81\% & 12.81\% & 2.01\% & \greybg{1.06\%} \\
vm1084 & 1,084 & 3.11\% & 5.98\% & 3.57\% & 6.79\% & 8.83\% & 5.18\% & 8.65\% & 1.95\% & \greybg{0.86\%} \\
pcb1173 & 1,173 & 3.97\% & 6.27\% & 3.79\% & 5.62\% & 6.89\% & 3.94\% & 3.51\% & 1.92\% & \greybg{1.35\%} \\
d1291 & 1,291 & 8.53\% & 13.46\% & 3.80\% & 11.03\% & 8.59\% & 9.95\% & 8.39\% & 6.20\% & \greybg{3.75\%} \\
rl1304 & 1,304 & 3.35\% & 7.75\% & 3.16\% & 14.60\% & 9.86\% & 8.15\% & 6.52\% & 4.67\% & \greybg{1.40\%} \\
rl1323 & 1,323 & 6.52\% & 9.27\% & 3.64\% & 10.20\% & 9.14\% & 9.34\% & 4.25\% & 2.66\% & \greybg{1.54\%} \\
nrw1379 & 1,379 & 5.07\% & 9.84\% & 3.01\% & 8.92\% & 4.84\% & 3.46\% & 3.30\% & 1.56\% & \greybg{1.22\%} \\
fl1400 & 1,400 & 16.40\% & 18.22\% & 24.22\% & 18.68\% & 25.23\% & 25.96\% & 321.08\% & 23.00\% & \greybg{10.14\%} \\
u1432 & 1,432 & 3.20\% & 3.32\% & 5.38\% & 5.74\% & 5.45\% & 4.44\% & 3.75\% & 1.51\% & \greybg{0.91\%} \\
fl1577 & 1,577 & 11.98\% & 17.69\% & 16.04\% & 17.22\% & 14.77\% & 9.49\% & 53.23\% & 13.01\% & \greybg{8.15\%} \\
d1655 & 1,655 & 11.92\% & 13.32\% & 14.00\% & 12.50\% & 14.29\% & 14.86\% & 15.50\% & 5.71\% & \greybg{3.52\%} \\
vm1748 & 1,748 & 7.70\% & 14.82\% & 7.36\% & 10.02\% & 8.02\% & 5.44\% & 11.39\% & 6.47\% & \greybg{1.29\%} \\
u1817 & 1,817 & 8.57\% & 9.31\% & 6.31\% & 12.02\% & 9.00\% & 8.51\% & 9.25\% & 5.95\% & \greybg{2.11\%} \\
rl1889 & 1,889 & 8.85\% & 10.65\% & 8.14\% & 15.50\% & 14.67\% & 12.48\% & 11.48\% & 7.08\% & \greybg{2.15\%} \\
d2103 & 2,103 & 12.27\% & 14.57\% & 10.40\% & \greybg{4.13\%} & 13.16\% & 8.13\% & 30.20\% & 21.05\% & 7.92\% \\
u2152 & 2,152 & 7.61\% & 10.58\% & 9.22\% & 10.69\% & 8.57\% & 9.10\% & 13.39\% & 5.37\% & \greybg{2.25\%} \\
u2319 & 2,319 & 2.14\% & 2.51\% & 1.40\% & 4.59\% & 0.80\% & 0.57\% & 3.40\% & 0.43\% & \greybg{0.33\%} \\
pr2392 & 2,392 & 8.99\% & 10.97\% & 7.00\% & 12.20\% & 11.99\% & 8.07\% & 11.65\% & 3.44\% & \greybg{1.88\%} \\
pcb3038 & 3,038 & 11.27\% & 12.97\% & 9.20\% & 13.68\% & 7.54\% & 6.36\% & 14.57\% & 2.61\% & \greybg{1.78\%} \\
fl3795 & 3,795 & 36.92\% & 13.89\% & 57.49\% & 21.20\% & 16.40\% & 23.77\% & 778.76\% & 20.39\% & \greybg{12.88\%} \\
fnl4461 & 4,461 & 21.14\% & 18.67\% & 26.67\% & 15.11\% & 7.93\% & 4.09\% & 56.00\% & 1.84\% & \greybg{1.76\%} \\
rl5915 & 5,915 & 21.94\% & 22.34\% & 44.49\% & 19.84\% & 11.92\% & 10.04\% & 146.46\% & 10.74\% & \greybg{2.92\%} \\
rl5934 & 5,934 & 25.11\% & 35.47\% & 48.62\% & 23.83\% & 13.23\% & 12.39\% & 163.71\% & 8.52\% & \greybg{3.59\%} \\
pla7397 & 7,397 & 28.99\% & 40.95\% & 89.03\% & 18.43\% & 10.01\% & 7.22\% & 544.94\% & 12.54\% & \greybg{4.26\%} \\
rl11849 & 11,849 & 43.12\% & 41.68\% & 116.23\% & 26.05\% & 12.62\% & 6.94\% & 545.65\% & 6.40\% & \greybg{2.96\%} \\
usa13509 & 13,509 & OOM & 55.79\% & 213.73\% & 27.37\% & 10.92\% & 6.99\% & 1440.84\% & 7.13\% & \greybg{2.54\%} \\
brd14051 & 14,051 & OOM & 42.47\% & 145.47\% & 24.66\% & 8.76\% & 6.45\% & 835.20\% & 3.92\% & \greybg{2.42\%} \\
d15112 & 15,112 & OOM & 34.40\% & 145.31\% & 24.80\% & 7.57\% & 7.53\% & 839.78\% & 2.21\% & \greybg{2.09\%} \\
d18512 & 18,512 & OOM & OOM & 136.29\% & 24.75\% & 7.31\% & 5.29\% & 780.01\% & 2.19\% & \greybg{2.07\%} \\
pla33810 & 33,810 & OOM & OOM & OOM & 29.74\% & 7.83\% & 6.38\% & 1695.12\% & 6.27\% & \greybg{4.40\%} \\
pla85900 & 85,900 & OOM & OOM & OOM & 26.14\% & 5.32\% & \greybg{4.22\%} & OOM & 5.11\% & 4.39\% \\
\end{longtable}
\endgroup

\clearpage
\begingroup
\footnotesize
\setlength{\tabcolsep}{1pt}
\renewcommand{\arraystretch}{1.04}
\setlength{\LTcapwidth}{\textwidth}
\begin{longtable}{@{}l|r|r|r|r|r|r|r|r|r|r@{}}
\caption{Detail results on CVRPLIB: Constructive NCO with Dynamic Embeddings.}\label{tab:cvrplib_dynamic}\\
\toprule[0.8pt]
\multicolumn{1}{c|}{Instance} & \multicolumn{1}{c|}{Scale} & \multicolumn{1}{c|}{\shortstack{BQ\\greedy}} & \multicolumn{1}{c|}{\shortstack{LEHD\\greedy}} & \multicolumn{1}{c|}{\shortstack{SIGD\\greedy}} & \multicolumn{1}{c|}{\shortstack{L2C-Insert\\greedy}} & \multicolumn{1}{c|}{\shortstack{INViT-3V\\greedy}} & \multicolumn{1}{c|}{\shortstack{L2R\\greedy}} & \multicolumn{1}{c|}{\shortstack{Rec-NCO\\greedy}} & \multicolumn{1}{c|}{\shortstack{MiLoop\\greedy}} & \multicolumn{1}{c}{\shortstack{MiLoop\\($M=100$)}} \\
\midrule[0.4pt]
\endfirsthead
\caption[]{Detail results on CVRPLIB: Constructive NCO with Dynamic Embeddings. (continued)}\\
\toprule[0.8pt]
\multicolumn{1}{c|}{Instance} & \multicolumn{1}{c|}{Scale} & \multicolumn{1}{c|}{\shortstack{BQ\\greedy}} & \multicolumn{1}{c|}{\shortstack{LEHD\\greedy}} & \multicolumn{1}{c|}{\shortstack{SIGD\\greedy}} & \multicolumn{1}{c|}{\shortstack{L2C-Insert\\greedy}} & \multicolumn{1}{c|}{\shortstack{INViT-3V\\greedy}} & \multicolumn{1}{c|}{\shortstack{L2R\\greedy}} & \multicolumn{1}{c|}{\shortstack{Rec-NCO\\greedy}} & \multicolumn{1}{c|}{\shortstack{MiLoop\\greedy}} & \multicolumn{1}{c}{\shortstack{MiLoop\\($M=100$)}} \\
\midrule[0.4pt]
\endhead
\midrule[0.4pt]
\multicolumn{11}{r}{\textit{Continued on the next page.}}\\
\endfoot
\bottomrule[0.8pt]
\endlastfoot
\multicolumn{11}{l}{\textit{Set-X}}\\*
X-n101-k25 & 100 & 9.62\% & 13.98\% & 13.85\% & 7.71\% & 12.24\% & 10.30\% & 20.96\% & 5.50\% & \greybg{4.22\%} \\
X-n106-k14 & 105 & 4.44\% & 3.73\% & \greybg{2.60\%} & 3.41\% & 4.97\% & 3.65\% & 5.17\% & 3.69\% & 2.74\% \\
X-n110-k13 & 109 & 4.86\% & 1.83\% & 2.33\% & 2.42\% & 7.63\% & 5.41\% & 3.99\% & 2.71\% & \greybg{0.81\%} \\
X-n115-k10 & 114 & 21.84\% & 9.39\% & 23.79\% & 11.80\% & 12.58\% & 18.22\% & 21.92\% & 3.58\% & \greybg{3.49\%} \\
X-n120-k6 & 119 & 3.19\% & 3.89\% & 2.60\% & 2.51\% & 12.83\% & 7.16\% & 2.75\% & 3.00\% & \greybg{1.04\%} \\
X-n125-k30 & 124 & 11.45\% & 17.88\% & 39.64\% & 5.43\% & 14.33\% & 6.77\% & 14.13\% & 11.46\% & \greybg{4.19\%} \\
X-n129-k18 & 128 & 3.26\% & 3.99\% & 4.80\% & 3.87\% & 8.75\% & 3.26\% & 2.47\% & 2.37\% & \greybg{1.33\%} \\
X-n134-k13 & 133 & 5.53\% & 8.94\% & 7.87\% & 11.76\% & 7.33\% & 9.10\% & 7.15\% & 2.12\% & \greybg{1.29\%} \\
X-n139-k10 & 138 & 5.19\% & 3.07\% & 2.60\% & 3.52\% & 9.66\% & 3.91\% & 1.82\% & 3.67\% & \greybg{0.77\%} \\
X-n143-k7 & 142 & 3.17\% & 14.00\% & 4.59\% & 6.97\% & 14.68\% & 7.85\% & 8.61\% & 2.54\% & \greybg{1.31\%} \\
X-n148-k46 & 147 & 20.11\% & 38.96\% & 54.06\% & 7.82\% & 18.91\% & 12.34\% & 22.19\% & 7.37\% & \greybg{5.82\%} \\
X-n153-k22 & 152 & 29.56\% & 28.90\% & 148.46\% & 12.60\% & 20.31\% & 22.60\% & 28.44\% & 14.75\% & \greybg{11.11\%} \\
X-n157-k13 & 156 & 3.79\% & 4.66\% & 6.27\% & 3.95\% & 14.64\% & 4.65\% & \greybg{1.17\%} & 2.68\% & 1.80\% \\
X-n162-k11 & 161 & 6.15\% & 3.44\% & 5.16\% & 6.53\% & 5.86\% & 4.99\% & 2.26\% & 1.99\% & \greybg{1.58\%} \\
X-n167-k10 & 166 & 4.38\% & 5.03\% & 2.97\% & 12.13\% & 9.14\% & 4.11\% & \greybg{1.88\%} & 3.77\% & 1.98\% \\
X-n172-k51 & 171 & 20.65\% & 33.28\% & 95.97\% & 7.01\% & 13.22\% & 11.27\% & 28.04\% & 6.39\% & \greybg{6.01\%} \\
X-n176-k26 & 175 & 20.13\% & 27.01\% & 101.13\% & 16.17\% & 23.88\% & 17.73\% & 25.26\% & 13.02\% & \greybg{9.82\%} \\
X-n181-k23 & 180 & \greybg{0.73\%} & 1.40\% & 1.75\% & 1.08\% & 15.73\% & 1.98\% & 1.12\% & 1.59\% & 1.00\% \\
X-n186-k15 & 185 & 4.62\% & 3.60\% & 3.90\% & 2.70\% & 8.21\% & 5.15\% & 2.00\% & 2.63\% & \greybg{1.61\%} \\
X-n190-k8 & 189 & 4.96\% & 5.39\% & 4.99\% & 3.68\% & 8.13\% & 3.63\% & \greybg{3.19\%} & 5.66\% & 3.98\% \\
X-n195-k51 & 194 & 19.39\% & 15.46\% & 66.96\% & 6.54\% & 11.91\% & 8.97\% & 39.36\% & 5.70\% & \greybg{2.98\%} \\
X-n200-k36 & 199 & 5.48\% & 10.25\% & 9.67\% & 8.81\% & 8.96\% & 5.85\% & 7.41\% & 6.07\% & \greybg{4.00\%} \\
X-n204-k19 & 203 & 6.76\% & 2.66\% & 2.56\% & 3.32\% & 8.58\% & 4.49\% & 4.03\% & 3.55\% & \greybg{2.05\%} \\
X-n209-k16 & 208 & 2.71\% & 3.86\% & 2.56\% & 3.16\% & 10.37\% & 5.70\% & 3.11\% & 2.87\% & \greybg{1.61\%} \\
X-n214-k11 & 213 & 3.93\% & 7.11\% & 5.66\% & 9.10\% & 13.55\% & 12.90\% & 4.19\% & 5.55\% & \greybg{3.66\%} \\
X-n219-k73 & 218 & 11.37\% & 15.51\% & 11.98\% & 2.76\% & 50.81\% & 2.82\% & 17.35\% & 1.74\% & \greybg{1.10\%} \\
X-n223-k34 & 222 & 5.43\% & 5.80\% & 5.23\% & 3.63\% & 13.17\% & 8.64\% & 11.72\% & 2.39\% & \greybg{2.21\%} \\
X-n228-k23 & 227 & 24.63\% & 14.38\% & 126.43\% & 14.67\% & 19.77\% & 10.31\% & 23.61\% & 7.42\% & \greybg{5.22\%} \\
X-n233-k16 & 232 & 5.74\% & 5.60\% & 7.77\% & 5.86\% & 12.81\% & 7.47\% & 9.19\% & 4.28\% & \greybg{2.83\%} \\
X-n237-k14 & 236 & 3.14\% & 4.05\% & 4.25\% & 3.51\% & 14.74\% & 6.74\% & 4.73\% & 2.59\% & \greybg{1.43\%} \\
X-n242-k48 & 241 & 5.01\% & 6.01\% & 11.46\% & 5.46\% & 13.33\% & 5.02\% & 7.57\% & 4.11\% & \greybg{2.90\%} \\
X-n247-k50 & 246 & 23.03\% & 44.88\% & 96.11\% & 19.78\% & 17.29\% & 18.15\% & 31.59\% & 16.20\% & \greybg{13.70\%} \\
X-n251-k28 & 250 & 3.13\% & \greybg{2.34\%} & 4.31\% & 4.25\% & 8.53\% & 6.03\% & 2.70\% & 3.89\% & 2.95\% \\
X-n256-k16 & 255 & 7.57\% & 7.26\% & 4.79\% & 12.03\% & 16.68\% & 8.09\% & 6.76\% & 3.85\% & \greybg{3.13\%} \\
X-n261-k13 & 260 & \greybg{2.92\%} & 8.78\% & 3.74\% & 7.99\% & 12.38\% & 9.34\% & 9.58\% & 5.03\% & 2.97\% \\
X-n266-k58 & 265 & 4.94\% & 6.38\% & 7.60\% & 5.80\% & 11.08\% & 7.61\% & 12.17\% & 4.97\% & \greybg{3.47\%} \\
X-n270-k35 & 269 & 5.78\% & 6.56\% & 5.67\% & 4.11\% & 10.61\% & 4.83\% & 2.70\% & 3.61\% & \greybg{2.68\%} \\
X-n275-k28 & 274 & 3.37\% & 6.54\% & 4.35\% & 3.39\% & 17.04\% & 4.06\% & 2.28\% & 4.99\% & \greybg{2.09\%} \\
X-n280-k17 & 279 & 8.93\% & 9.64\% & 15.76\% & 6.98\% & 12.87\% & 7.57\% & 16.42\% & 4.84\% & \greybg{3.46\%} \\
X-n284-k15 & 283 & 7.27\% & 4.94\% & 5.43\% & \greybg{4.54\%} & 10.44\% & 6.16\% & 5.73\% & 6.08\% & 4.83\% \\
X-n289-k60 & 288 & 7.92\% & 9.88\% & 16.16\% & 6.79\% & 11.18\% & 8.13\% & 16.61\% & 7.40\% & \greybg{5.98\%} \\
X-n294-k50 & 293 & 10.19\% & 9.67\% & 13.37\% & 6.25\% & 14.90\% & 6.45\% & 18.84\% & 4.10\% & \greybg{2.96\%} \\
X-n298-k31 & 297 & 5.03\% & 7.60\% & 5.42\% & 6.29\% & 12.38\% & 9.68\% & 3.64\% & 3.13\% & \greybg{1.93\%} \\
X-n303-k21 & 302 & 11.22\% & \greybg{2.78\%} & 6.51\% & 4.14\% & 11.85\% & 4.73\% & 11.17\% & 3.19\% & 2.95\% \\
X-n308-k13 & 307 & 10.78\% & 4.42\% & 11.86\% & 8.34\% & 12.26\% & 6.79\% & 15.03\% & 5.70\% & \greybg{3.50\%} \\
X-n313-k71 & 312 & 10.82\% & 14.05\% & 19.75\% & 5.81\% & 8.41\% & 7.13\% & 20.52\% & 5.16\% & \greybg{4.55\%} \\
X-n317-k53 & 316 & 2.69\% & 3.86\% & 4.10\% & 2.13\% & 21.86\% & \greybg{1.47\%} & 16.19\% & 2.20\% & 2.05\% \\
X-n322-k28 & 321 & 6.22\% & 3.83\% & 4.31\% & 7.39\% & 14.68\% & 7.21\% & 7.17\% & 4.01\% & \greybg{3.07\%} \\
X-n327-k20 & 326 & 4.34\% & 10.71\% & 4.93\% & 5.44\% & 9.53\% & 11.64\% & 4.98\% & 3.36\% & \greybg{3.02\%} \\
X-n331-k15 & 330 & 5.85\% & 3.33\% & 4.35\% & 3.61\% & 11.22\% & 5.92\% & 4.67\% & 4.03\% & \greybg{3.02\%} \\
X-n336-k84 & 335 & 12.24\% & 18.78\% & 43.43\% & 8.70\% & 11.10\% & 9.86\% & 19.96\% & 11.16\% & \greybg{8.58\%} \\
X-n344-k43 & 343 & 5.46\% & 3.90\% & 6.70\% & 4.86\% & 12.29\% & 7.56\% & 6.48\% & 3.87\% & \greybg{3.17\%} \\
X-n351-k40 & 350 & 8.92\% & 7.43\% & 6.71\% & 10.36\% & 13.08\% & 12.04\% & 10.38\% & 5.16\% & \greybg{3.32\%} \\
X-n359-k29 & 358 & \greybg{2.00\%} & 2.78\% & 4.62\% & 5.47\% & 11.27\% & 4.88\% & 7.04\% & 2.07\% & 2.07\% \\
X-n367-k17 & 366 & 8.69\% & 9.35\% & 15.82\% & 10.88\% & 14.61\% & 10.38\% & 15.02\% & 8.18\% & \greybg{6.40\%} \\
X-n376-k94 & 375 & 3.54\% & 5.45\% & 4.02\% & 2.45\% & 35.07\% & 2.62\% & 24.78\% & 2.05\% & \greybg{1.33\%} \\
X-n384-k52 & 383 & 5.50\% & 7.78\% & 4.90\% & 5.23\% & 8.71\% & 7.06\% & 7.40\% & 3.87\% & \greybg{2.52\%} \\
X-n393-k38 & 392 & 4.15\% & 4.45\% & 6.62\% & 5.15\% & 14.02\% & 7.56\% & 7.58\% & 4.19\% & \greybg{2.91\%} \\
X-n401-k29 & 400 & 7.65\% & 5.54\% & 4.52\% & 4.61\% & 9.69\% & 7.79\% & 9.75\% & 6.18\% & \greybg{2.99\%} \\
X-n411-k19 & 410 & 5.84\% & 13.40\% & 25.05\% & 13.37\% & 19.03\% & 16.37\% & 19.62\% & 6.98\% & \greybg{5.40\%} \\
X-n420-k130 & 419 & 19.83\% & 61.63\% & 66.91\% & 9.88\% & 19.16\% & 9.53\% & 32.49\% & 8.66\% & \greybg{7.69\%} \\
X-n429-k61 & 428 & 5.75\% & 6.24\% & 6.56\% & 4.01\% & 9.80\% & 7.39\% & 9.81\% & 3.59\% & \greybg{3.32\%} \\
X-n439-k37 & 438 & 3.36\% & \greybg{1.97\%} & 5.27\% & 2.76\% & 18.59\% & 4.56\% & 3.29\% & 2.76\% & 2.16\% \\
X-n449-k29 & 448 & 7.11\% & 7.17\% & 4.84\% & 4.90\% & 8.55\% & 9.70\% & 10.48\% & 4.21\% & \greybg{3.36\%} \\
X-n459-k26 & 458 & 10.65\% & 8.60\% & 9.16\% & 9.50\% & 14.42\% & 6.37\% & 11.28\% & 6.12\% & \greybg{4.61\%} \\
X-n469-k138 & 468 & 15.47\% & 28.32\% & 17.25\% & 8.60\% & 8.50\% & 8.28\% & 47.23\% & 7.31\% & \greybg{6.28\%} \\
X-n480-k70 & 479 & 4.58\% & \greybg{3.05\%} & 5.74\% & 5.20\% & 9.29\% & 5.93\% & 14.05\% & 4.33\% & 3.64\% \\
X-n491-k59 & 490 & 10.92\% & 5.06\% & 7.50\% & 5.54\% & 10.35\% & 7.39\% & 10.77\% & 5.05\% & \greybg{3.33\%} \\
X-n502-k39 & 501 & \greybg{2.40\%} & 3.26\% & 3.26\% & 4.21\% & 13.45\% & 3.00\% & 7.02\% & 5.77\% & 3.14\% \\
X-n513-k21 & 512 & 8.33\% & 4.90\% & 9.46\% & 8.90\% & 20.63\% & 9.74\% & 8.30\% & 5.30\% & \greybg{4.80\%} \\
X-n524-k153 & 523 & 32.41\% & 78.65\% & 83.71\% & 17.66\% & 17.15\% & 15.57\% & 36.96\% & 17.01\% & \greybg{13.73\%} \\
X-n536-k96 & 535 & 10.43\% & 13.89\% & 13.20\% & 6.14\% & 10.12\% & 6.57\% & 30.04\% & 5.47\% & \greybg{4.53\%} \\
X-n548-k50 & 547 & \greybg{1.45\%} & 3.31\% & 2.54\% & 2.58\% & 16.26\% & 4.97\% & 3.78\% & 2.55\% & 1.69\% \\
X-n561-k42 & 560 & 4.99\% & 5.82\% & 6.79\% & 4.65\% & 11.70\% & 7.61\% & 12.61\% & 3.54\% & \greybg{3.33\%} \\
X-n573-k30 & 572 & 14.33\% & 11.31\% & 18.91\% & 12.71\% & 12.03\% & 12.07\% & 11.22\% & 8.85\% & \greybg{6.82\%} \\
X-n586-k159 & 585 & 10.68\% & 17.10\% & 16.75\% & 7.78\% & 10.76\% & 6.05\% & 33.06\% & 7.06\% & \greybg{5.81\%} \\
X-n599-k92 & 598 & 4.49\% & 8.90\% & 6.05\% & 5.57\% & 8.79\% & 8.35\% & 16.98\% & 4.75\% & \greybg{4.12\%} \\
X-n613-k62 & 612 & 12.09\% & 6.86\% & 7.11\% & 6.15\% & 14.27\% & 9.89\% & 14.29\% & 5.15\% & \greybg{3.62\%} \\
X-n627-k43 & 626 & 4.80\% & 5.45\% & 6.75\% & \greybg{4.53\%} & 10.18\% & 7.25\% & 7.57\% & 19.75\% & 6.87\% \\
X-n641-k35 & 640 & 3.72\% & 5.64\% & 5.32\% & 6.36\% & 8.82\% & 8.19\% & 6.96\% & 4.61\% & \greybg{3.23\%} \\
X-n655-k131 & 654 & 3.98\% & 5.22\% & 4.51\% & 1.78\% & 27.00\% & 1.55\% & 51.93\% & 1.93\% & \greybg{1.43\%} \\
X-n670-k130 & 669 & 30.13\% & 98.94\% & 103.12\% & 18.59\% & 19.96\% & 17.49\% & 41.07\% & 19.34\% & \greybg{16.31\%} \\
X-n685-k75 & 684 & 12.97\% & 8.32\% & 11.74\% & 5.87\% & 14.44\% & 12.38\% & 18.35\% & 5.74\% & \greybg{3.69\%} \\
X-n701-k44 & 700 & 3.64\% & 4.10\% & 4.75\% & 6.49\% & 11.76\% & 7.86\% & 10.24\% & 3.60\% & \greybg{3.28\%} \\
X-n716-k35 & 715 & 20.82\% & 8.73\% & 6.55\% & 12.19\% & 10.67\% & 7.38\% & 18.14\% & 9.19\% & \greybg{6.00\%} \\
X-n733-k159 & 732 & 11.17\% & 14.70\% & 21.06\% & 6.27\% & 14.10\% & 8.48\% & 32.94\% & 4.50\% & \greybg{3.76\%} \\
X-n749-k98 & 748 & 11.07\% & 8.94\% & 10.32\% & 6.93\% & 12.21\% & 9.72\% & 31.93\% & 4.92\% & \greybg{4.22\%} \\
X-n766-k71 & 765 & 12.22\% & 15.48\% & 85.11\% & 10.47\% & 15.89\% & 12.42\% & 24.96\% & 10.98\% & \greybg{6.71\%} \\
X-n783-k48 & 782 & 13.95\% & 5.61\% & 5.82\% & 7.40\% & 8.50\% & 9.79\% & 12.65\% & 4.67\% & \greybg{3.64\%} \\
X-n801-k40 & 800 & \greybg{3.06\%} & 3.44\% & 5.97\% & 5.46\% & 15.27\% & 9.88\% & 7.84\% & 4.44\% & 3.50\% \\
X-n819-k171 & 818 & 12.12\% & 12.20\% & 12.78\% & 7.00\% & 9.09\% & 7.83\% & 53.49\% & 5.71\% & \greybg{5.06\%} \\
X-n837-k142 & 836 & \greybg{3.78\%} & 6.65\% & 7.57\% & 5.79\% & 10.01\% & 6.80\% & 33.07\% & 4.91\% & 4.14\% \\
X-n856-k95 & 855 & 3.43\% & 2.92\% & 5.29\% & \greybg{2.19\%} & 21.10\% & 5.95\% & 22.62\% & 2.94\% & 2.86\% \\
X-n876-k59 & 875 & 12.74\% & 7.27\% & 6.82\% & 7.16\% & 10.08\% & 7.52\% & 21.00\% & 7.81\% & \greybg{6.12\%} \\
X-n895-k37 & 894 & 5.80\% & 8.64\% & 8.37\% & 8.20\% & 13.23\% & 11.13\% & 13.20\% & 4.98\% & \greybg{4.18\%} \\
X-n916-k207 & 915 & 9.46\% & 15.36\% & 10.93\% & 6.52\% & 10.21\% & 6.64\% & 48.16\% & 6.32\% & \greybg{4.97\%} \\
X-n936-k151 & 935 & 32.42\% & 139.40\% & 139.39\% & 19.35\% & 21.45\% & 21.67\% & 54.97\% & 19.83\% & \greybg{17.59\%} \\
X-n957-k87 & 956 & 3.59\% & 3.81\% & 5.26\% & 3.82\% & 16.48\% & 6.60\% & 16.00\% & 3.05\% & \greybg{2.46\%} \\
X-n979-k58 & 978 & 10.71\% & 19.75\% & 7.84\% & 8.40\% & 9.36\% & 8.13\% & 36.32\% & 23.17\% & \greybg{7.66\%} \\
X-n1001-k43 & 1,000 & \greybg{3.51\%} & 7.53\% & 7.97\% & 7.51\% & 12.57\% & 10.29\% & 13.33\% & 4.58\% & 4.57\% \\
\midrule[0.4pt]
\multicolumn{11}{l}{\textit{Set-AGS}}\\*
Leuven1 & 3,000 & 21.06\% & 16.36\% & 17.01\% & 7.07\% & 13.66\% & 11.42\% & 75.11\% & 4.99\% & \greybg{4.29\%} \\
Leuven2 & 4,000 & 38.91\% & 33.07\% & 25.12\% & 25.72\% & 26.40\% & 12.53\% & 248.74\% & 13.60\% & \greybg{12.23\%} \\
Antwerp1 & 6,000 & 19.62\% & 15.16\% & 13.30\% & 6.23\% & 15.37\% & 10.99\% & 146.99\% & 7.38\% & \greybg{5.14\%} \\
Antwerp2 & 7,000 & 32.30\% & 21.33\% & 17.38\% & 19.96\% & 27.66\% & 12.84\% & 370.07\% & 12.22\% & \greybg{10.28\%} \\
Ghent1 & 10,000 & 47.79\% & 27.63\% & 24.03\% & 11.14\% & 15.76\% & 10.03\% & 339.67\% & 5.89\% & \greybg{4.89\%} \\
Ghent2 & 11,000 & 58.76\% & 39.79\% & 29.48\% & 30.28\% & 30.63\% & 11.04\% & 1187.88\% & 13.14\% & \greybg{10.57\%} \\
Brussels1 & 15,000 & OOM & OOM & 24.66\% & 15.35\% & 17.94\% & 12.49\% & 556.19\% & 8.35\% & \greybg{6.94\%} \\
Brussels2 & 16,000 & OOM & OOM & 36.01\% & 29.63\% & 31.86\% & 12.02\% & 1111.06\% & 13.37\% & \greybg{11.41\%} \\
Flanders1 & 20,000 & OOM & OOM & 45.69\% & 18.05\% & 23.41\% & 7.84\% & 697.45\% & 10.69\% & \greybg{6.78\%} \\
Flanders2 & 30,000 & OOM & OOM & 133.60\% & 45.56\% & 39.60\% & 13.07\% & 2668.36\% & 12.90\% & \greybg{12.24\%} \\
\end{longtable}
\endgroup
\clearpage

\section{Visualization}
\subsection{Visualization on Synthetic Dataset}
\label{append:synthetic_visualization}

\begin{figure}[H]
    \centering
    \includegraphics[width=\linewidth]{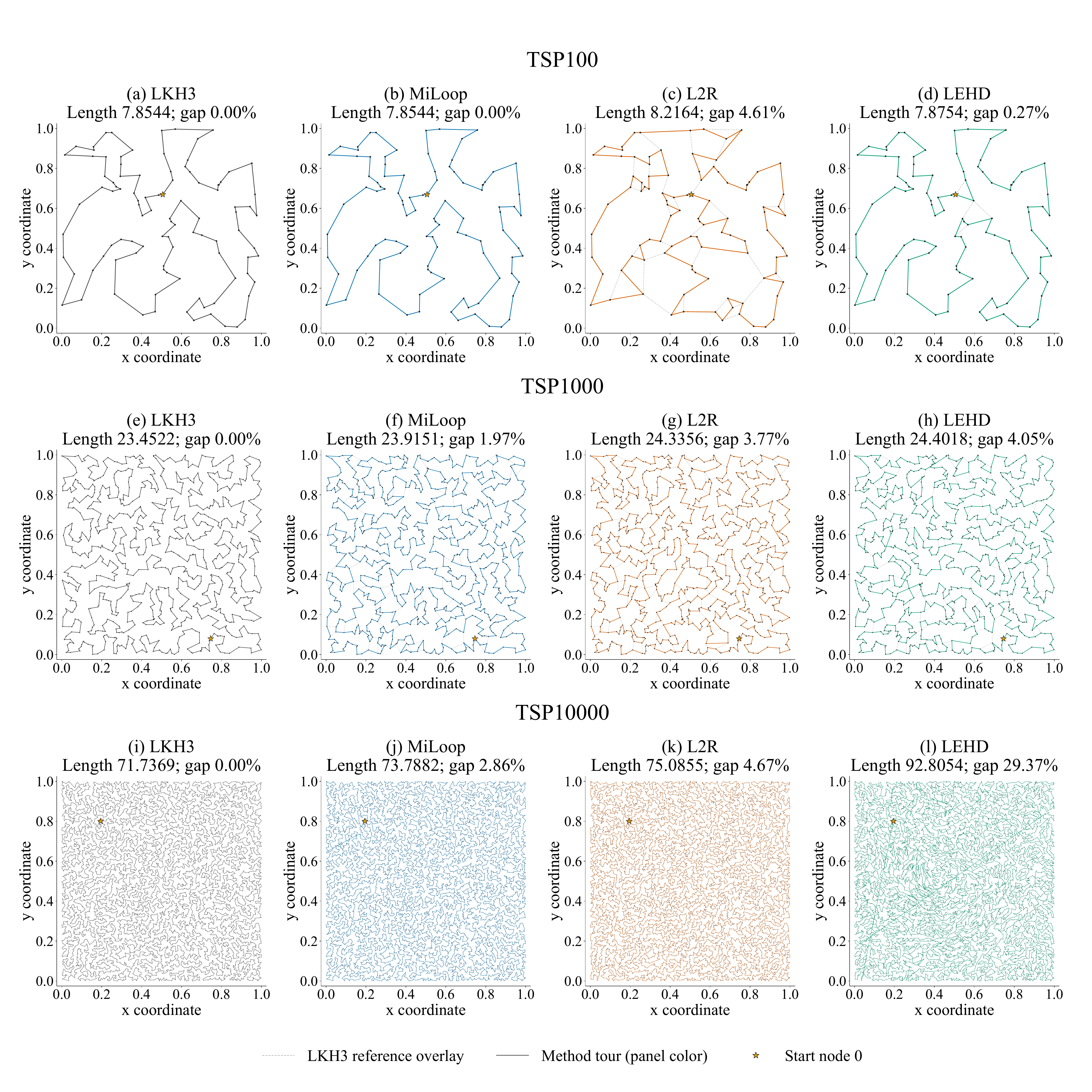}
    \caption{Solution visualizations on synthetic TSP instances. Rows show TSP100, TSP1000, and TSP10000, while columns show LKH3, MiLoop, L2R, and LEHD, respectively. All methods solve the same instance within each row. Gray dashed lines overlay the LKH3 reference tour, and stars mark the starting node. Each panel reports the tour length and its gap relative to the LKH3 solution. All results are obtained via greedy decoding.}
    \label{fig:synthetic_tsp_visualization}
\end{figure}

\clearpage
\begin{figure}[H]
    \centering
    \includegraphics[width=\linewidth]{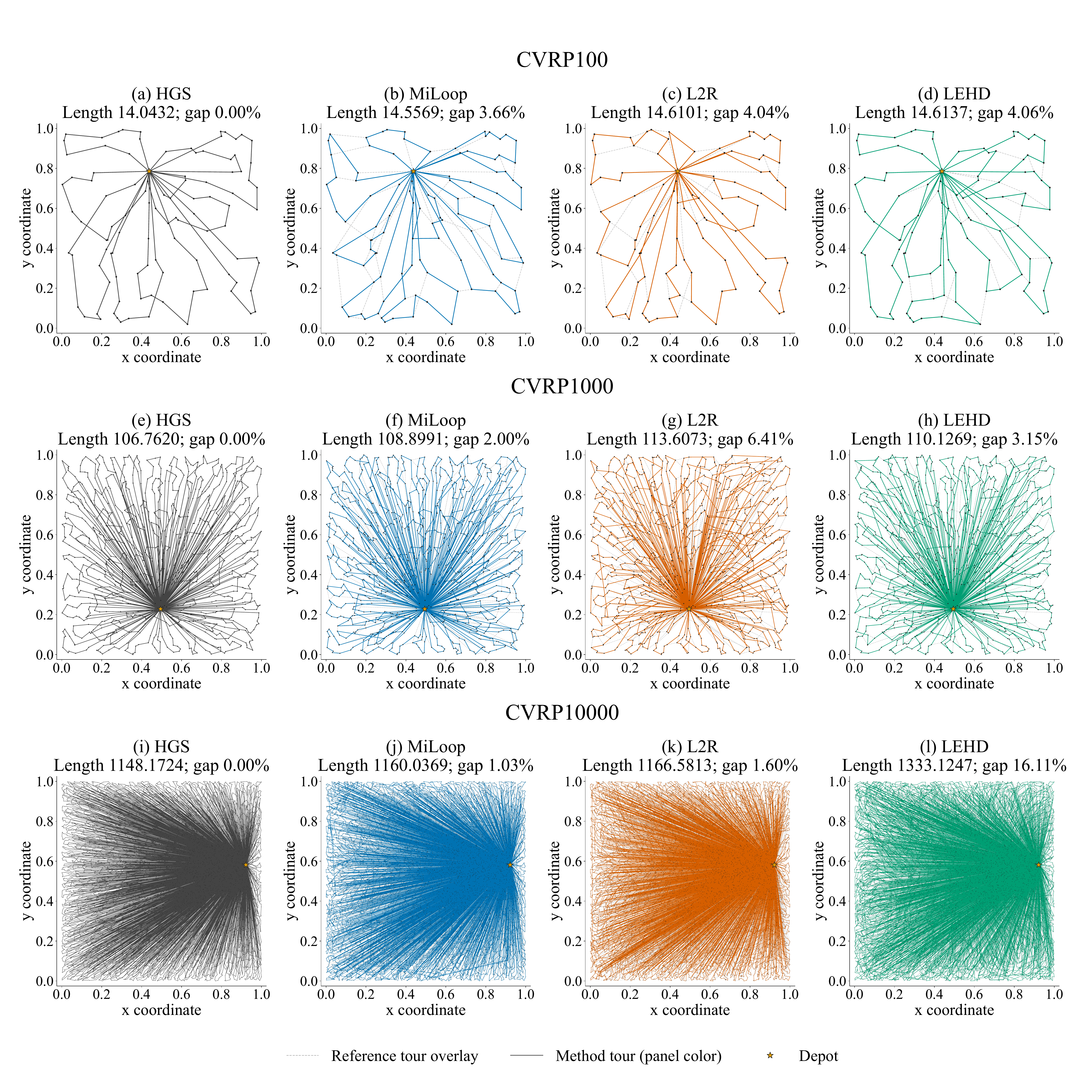}
    \caption{Solution visualizations on synthetic CVRP instances. Rows show CVRP100, CVRP1000, and CVRP10000. The first column shows the reference solutions, obtained by HGS; the remaining columns show MiLoop, L2R, and LEHD, respectively. All methods solve the same instance within each row. Gray dashed lines overlay the reference routes, and stars mark the depot. Each panel reports the total route length and its gap relative to the reference solution. All results are obtained via greedy decoding.}
    \label{fig:synthetic_cvrp_visualization}
\end{figure}

\clearpage

\subsection{Visualization on Benchmark Dataset}
\label{append:benchmark_visualization}

\begin{figure}[H]
    \centering
    \captionsetup[subfigure]{font=footnotesize,skip=2pt,justification=centering}
    \begin{subfigure}[t]{0.43\linewidth}
        \centering
        \includegraphics[width=\linewidth,trim=0 30bp 0 0,clip]{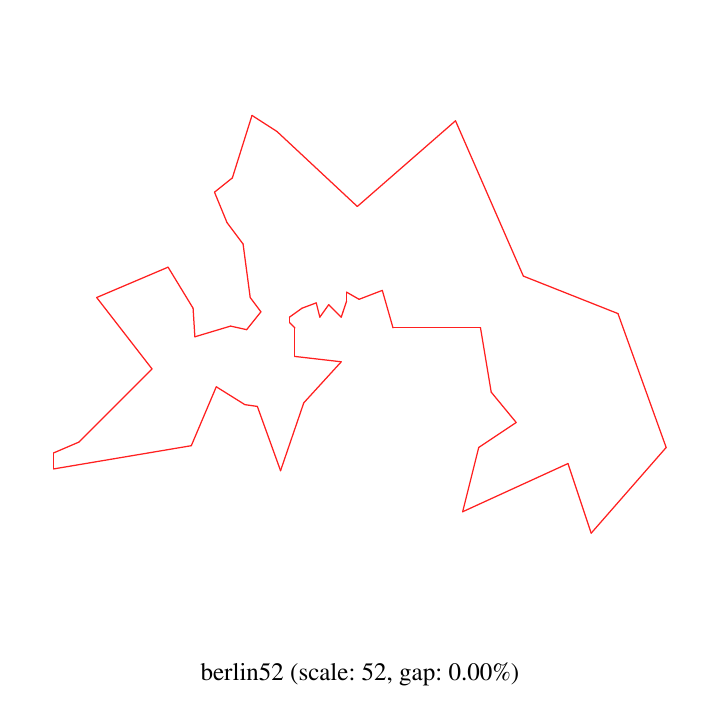}
        \caption{berlin52 ($N=52$, gap $0.00\%$)}
        \label{fig:lib_berlin52}
    \end{subfigure}\hspace{0.03\linewidth}%
    \begin{subfigure}[t]{0.43\linewidth}
        \centering
        \includegraphics[width=\linewidth,trim=0 30bp 0 0,clip]{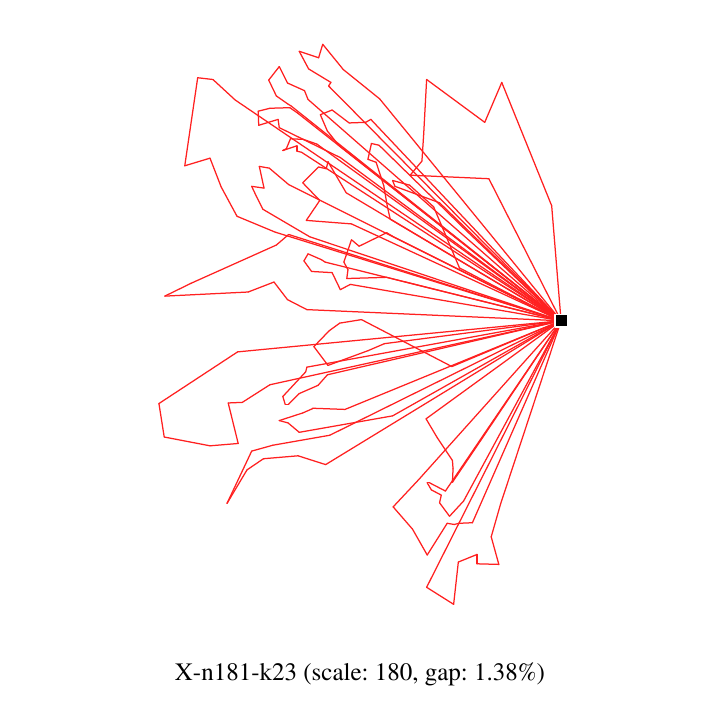}
        \caption{X-n181-k23 ($N=180$, gap $1.38\%$)}
        \label{fig:lib_x_n181_k23}
    \end{subfigure}

    \par\vspace{4pt}
    \begin{subfigure}[t]{0.43\linewidth}
        \centering
        \includegraphics[width=\linewidth,trim=0 30bp 0 0,clip]{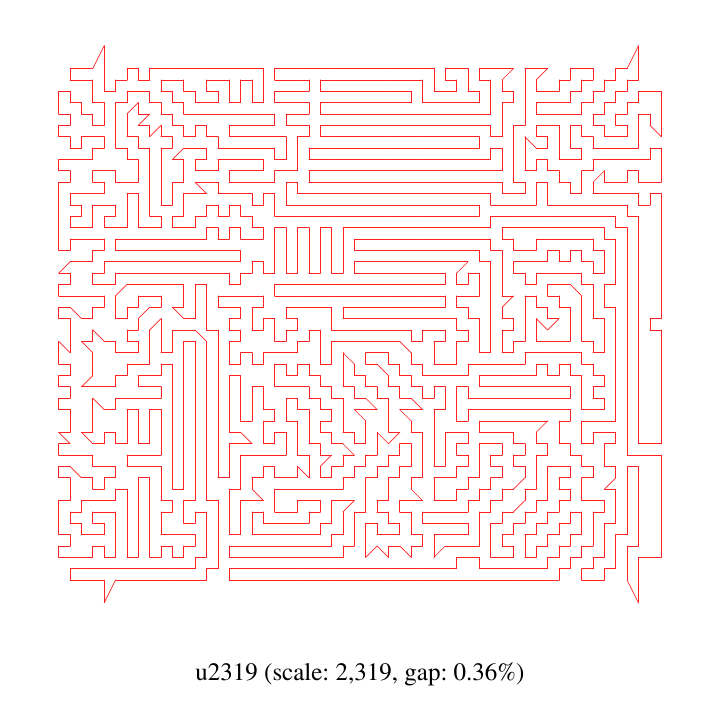}
        \caption{u2319 ($N=2{,}319$, gap $0.36\%$)}
        \label{fig:lib_u2319}
    \end{subfigure}\hspace{0.03\linewidth}%
    \begin{subfigure}[t]{0.43\linewidth}
        \centering
        \includegraphics[width=\linewidth,trim=0 30bp 0 0,clip]{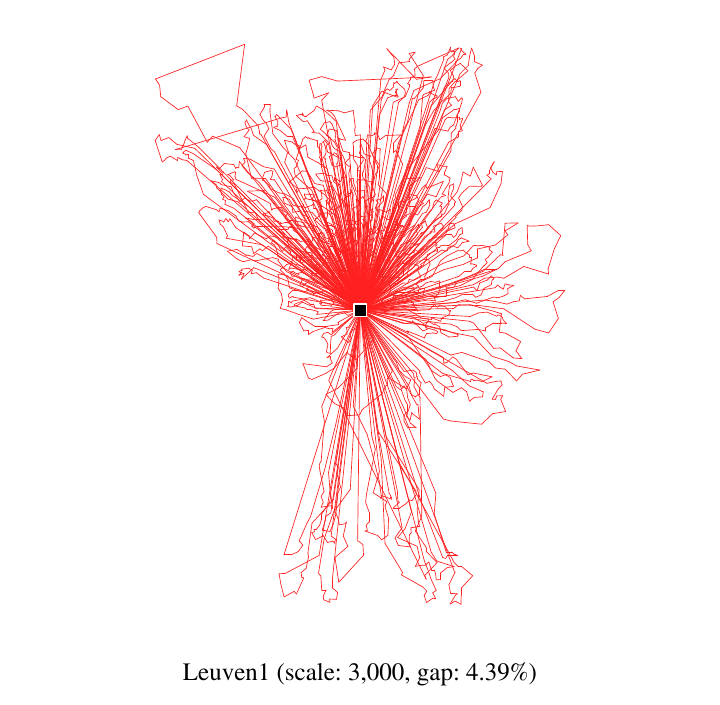}
        \caption{Leuven1 ($N=3{,}000$, gap $4.39\%$)}
        \label{fig:lib_leuven1}
    \end{subfigure}

    \par\vspace{4pt}
    \begin{subfigure}[t]{0.43\linewidth}
        \centering
        \includegraphics[width=\linewidth,trim=0 30bp 0 0,clip]{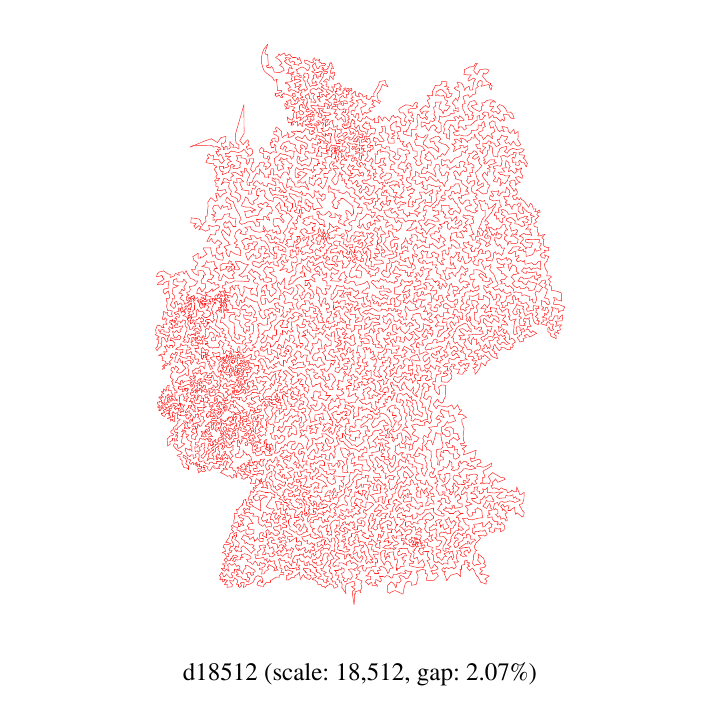}
        \caption{d18512 ($N=18{,}512$, gap $2.07\%$)}
        \label{fig:lib_d18512}
    \end{subfigure}\hspace{0.03\linewidth}%
    \begin{subfigure}[t]{0.43\linewidth}
        \centering
        \includegraphics[width=\linewidth,trim=0 30bp 0 0,clip]{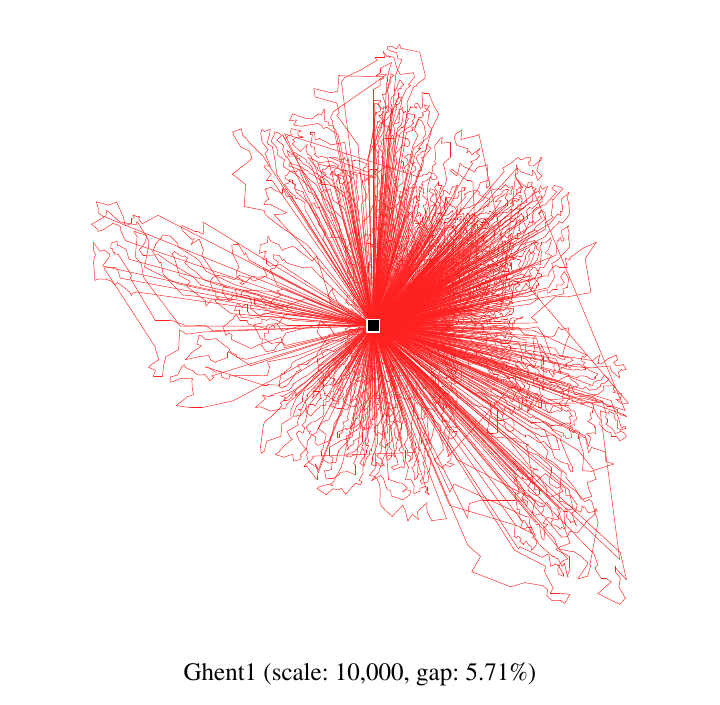}
        \caption{Ghent1 ($N=10{,}000$, gap $5.71\%$)}
        \label{fig:lib_ghent1}
    \end{subfigure}
    \caption{Solution visualizations on TSPLib (left) and CVRPLib (right), ordered by increasing instance size within each column. Red lines show solution edges, and black squares mark CVRP depots. Instance names, sizes, and gaps are given in the subcaptions. All results are obtained via greedy decoding.}
    \label{fig:benchmark_visualization}
\end{figure}
\clearpage

\section{Licenses}
\label{append:licenses}

\begin{table}[htbp]
    \centering
    \caption{Licenses and stated usage terms of baseline code and benchmark datasets.}
    \label{tab:resource_licenses}
    \begin{threeparttable}
    \fontsize{8}{9.5}\selectfont
    \setlength{\tabcolsep}{3pt}
    \renewcommand{\arraystretch}{1.18}
    \urlstyle{same}
    \begin{tabularx}{\linewidth}{@{}l l >{\raggedright\arraybackslash}p{0.47\linewidth} >{\raggedright\arraybackslash}X@{}}
    \toprule
    Resource & Type & Link & License / usage terms \\
    \midrule
    Concorde & Code & \url{https://www.math.uwaterloo.ca/tsp/concorde.html} & \href{https://www.math.uwaterloo.ca/tsp/concorde.html}{Academic research use} \\
    OR-Tools & Code & \url{https://github.com/google/or-tools} & \href{https://github.com/google/or-tools/blob/stable/LICENSE}{Apache-2.0} \\
    LKH3 & Code & \url{http://webhotel4.ruc.dk/~keld/research/LKH-3/} & \href{http://webhotel4.ruc.dk/~keld/research/LKH-3/}{Academic and non-commercial use} \\
    HGS & Code & \url{https://github.com/vidalt/HGS-CVRP} & \href{https://github.com/vidalt/HGS-CVRP/blob/main/LICENSE}{MIT} \\
    POMO & Code & \url{https://github.com/yd-kwon/POMO} & \href{https://github.com/yd-kwon/POMO/blob/master/NEW_py_ver/License.md}{MIT} \\
    Omni-VRP & Code & \url{https://github.com/RoyalSkye/Omni-VRP} & \href{https://github.com/RoyalSkye/Omni-VRP/blob/main/LICENSE}{MIT} \\
    ELG & Code & \url{https://github.com/gaocrr/ELG} & \href{https://github.com/gaocrr/ELG/blob/master/LICENSE}{MIT} \\
    ICAM & Code & \url{https://github.com/CIAM-Group/ICAM} & \href{https://github.com/CIAM-Group/ICAM/blob/main/LICENSE}{MIT} \\
    Heter-AM & Code & \url{https://github.com/jingwenli0312/Heterogeneous-Attentions-PDP-DRL} & \href{https://github.com/jingwenli0312/Heterogeneous-Attentions-PDP-DRL/blob/main/LICENSE}{MIT} \\
    URS & Code & \url{https://github.com/CIAM-Group/URS} & \href{https://github.com/CIAM-Group/URS/blob/main/LICENSE}{AGPL-3.0}\\
    BQ & Code & \url{https://github.com/naver/bq-nco} & \href{https://github.com/naver/bq-nco/blob/main/LICENCE}{CC BY-NC-SA 4.0} \\
    LEHD & Code & \href{https://github.com/CIAM-Group/NCO_code/tree/main/single_objective/LEHD}{\nolinkurl{github.com/CIAM-Group/NCO_code} (LEHD)} & \href{https://github.com/CIAM-Group/NCO_code/blob/main/LICENSE}{MIT} \\
    SIGD & Code & \url{https://github.com/grimmlab/gumbeldore} & \href{https://github.com/grimmlab/gumbeldore/blob/main/LICENSE}{MIT} \\
    L2C-Insert & Code & \url{https://github.com/CIAM-Group/L2C_Insert} & \href{https://github.com/CIAM-Group/L2C_Insert/blob/main/LICENSE}{MIT} \\
    INViT & Code & \url{https://github.com/Kasumigaoka-Utaha/INViT} & Available for academic research use \\
    L2R & Code & \url{https://github.com/CIAM-Group/L2R} & \href{https://github.com/CIAM-Group/L2R/blob/main/LICENSE}{MIT} \\
    Rec-NCO & Code & \url{https://github.com/TimD3/Recurrent-NCO} & \href{https://github.com/TimD3/Recurrent-NCO/blob/main/LICENSE}{CC BY-NC-SA 4.0} \\
    H-TSP & Code & \url{https://github.com/Learning4Optimization-HUST/H-TSP} & \href{https://github.com/Learning4Optimization-HUST/H-TSP/blob/main/LICENSE}{MIT} \\
    GLOP & Code & \url{https://github.com/henry-yeh/GLOP} & \href{https://github.com/henry-yeh/GLOP/blob/master/Licence}{MIT} \\
    UDC & Code & \href{https://github.com/CIAM-Group/NCO_code/tree/main/single_objective/UDC-Large-scale-CO-master}{\nolinkurl{github.com/CIAM-Group/NCO_code} (UDC)} & \href{https://github.com/CIAM-Group/NCO_code/blob/main/LICENSE}{MIT} \\
    \midrule
    TSPLib & Dataset & \url{https://comopt.ifi.uni-heidelberg.de/software/TSPLIB95/} & Available for academic research use \\
    CVRPLib & Dataset & \url{https://galgos.inf.puc-rio.br/cvrplib/} & \href{https://galgos.inf.puc-rio.br/cvrplib/index.php/en/register/terms}{Research use} \\
    \bottomrule
    \end{tabularx}
    \end{threeparttable}
\end{table}

\end{CJK*}
\end{document}